\documentclass{article}
\usepackage[table]{xcolor}
\usepackage[utf8]{inputenc}
\usepackage{authblk}
\usepackage{setspace}
\usepackage{graphicx}
\usepackage{subcaption}
\usepackage{amsmath}
\usepackage{lineno}
\usepackage{hyperref}
\usepackage[top=2cm, bottom=2cm, left=1.5cm, right=1.5cm, margin=1.25in]{geometry}
\usepackage{tikz}
\usepackage{longtable,array}
\usepackage{colortbl}
\usepackage{makecell}
\definecolor{domainrow}{RGB}{230, 230, 230}
\definecolor{ourrow}{RGB}{230, 240, 255}

\usepackage{amssymb}      
\usepackage{pifont}       

\newcommand{\greencheck}{{\color{green!60!black}\checkmark}} 
\newcommand{\redcross}{{\color{red!70!black}\ding{55}}}       
\usepackage{pifont} 
\usepackage{tabularx, booktabs, multirow}
\usetikzlibrary{positioning, shapes.geometric, backgrounds}

\usepackage[style=nejm, 
citestyle=numeric-comp,
sorting=none]{biblatex}
\title{Prompt Engineering in the Segment Anything Model: Methodologies, Applications, and Emerging Challenges}

\author[1]{Yidong Jiang}
\author[1,*]{Jiangtong Li}
\author[1]{Dawei Cheng}

\affil[1]{School of Computer Science and Technology, Tongji University, Shanghai, China.}

\affil[*]{Address correspondence to: jiangtongli@tongji.edu.cn}

\date{}

\begin{document}

\maketitle

\begin{abstract}
The Segment Anything Model (SAM) has transformed image segmentation by introducing a prompt-based paradigm that enables strong zero-shot generalization. In this framework, prompts serve as a semantic interface between human intent and machine perception, making prompt engineering a central factor in model performance. Despite its importance, prompt engineering within SAM and its variants has not yet been systematically reviewed in the literature. This survey addresses that gap by providing a structured and comprehensive overview of prompt engineering techniques developed for SAM and its rapidly growing ecosystem. We introduce a hierarchical taxonomy that organizes methods into geometric prompts, textual semantic prompts, and multimodal fusion prompts, and analyze how these categories reflect different design principles and application goals. In addition, we examine the transition from manually crafted prompts to more advanced, automated approaches based on detector outputs, prototype learning, reinforcement learning, and vision–language models. Beyond categorizing existing work, we trace how prompt engineering has enabled SAM to generalize across domains such as medical imaging, remote sensing, industrial inspection, and anomaly detection. We further identify key challenges—including prompt sensitivity, cross-modal misalignment, and computational inefficiency—and highlight promising research directions such as causal prompt reasoning, collaborative multi-agent prompting, and diffusion-based progressive refinement. By consolidating these developments into a unified perspective, our survey provides a timely reference for understanding the role of prompt engineering in segmentation foundation models and lays the groundwork for future advances in this evolving field.
\end{abstract}


\section{Introduction}
Image segmentation, as a fundamental task in computer vision, has long faced a trade-off between generalization and task-specific adaptation. Traditional segmentation methods often require large amounts of task-specific training data and typically struggle to generalize to novel scenarios without extensive fine-tuning. Moreover, they lack an intuitive mechanism for users to explicitly convey segmentation intent, limiting their flexibility and precision in diverse real-world applications. The emergence of the Segment Anything Model (SAM)~\cite{kirillov2023segment} fundamentally reshapes this landscape through its innovative promptable framework, enabling remarkable zero-shot generalization. In this paradigm, user-provided prompts serve as the critical semantic bridge between human intent and machine understanding. Rather than acting as mere auxiliary inputs, prompts play a central role in guiding the segmentation process, enabling SAM to dynamically modulate its focus and achieve remarkable zero-shot generalization without additional training. As a result, the model’s segmentation precision and cross-domain adaptability are closely tied to the design and quality of the input prompts, positioning prompt engineering as a foundational mechanism underlying SAM’s performance.

\begin{table*}[ht]
\centering
\caption{Comparison of representative SAM-related surveys. Our work is distinguished by its exclusive focus on prompt engineering, the introduction of a structured and unified taxonomy, and a systematic analysis of prompt engineering across different prompting paradigms.}

\label{tab:sam_survey_comparison}
\small
\renewcommand{\arraystretch}{1.15}
\begin{tabular}{p{3.8cm}ccc}
\toprule
\textbf{Survey} & \textbf{Prompt Eng. Focus} & \textbf{Prompt Taxonomy} & \textbf{Systematic Analysis} \\
\midrule
Zhang \emph{et al.} (2023a)~\cite{zhang2023comprehensivesurveysegmentmodel} & \redcross & \redcross & \redcross \\
Zhang \emph{et al.} (2023b)~\cite{10386032} & \redcross & \redcross & \redcross \\
Zhang \emph{et al.} (2023c)~\cite{10386032} & \redcross & \redcross & \greencheck \\
Zhang \emph{et al.} (2024)~\cite{zhang2024surveysegmentmodelsam} & \redcross & \redcross & \greencheck \\
Zhang \emph{et al.} (2025)~\cite{jiaxing2025sam2imagevideosegmentation} & \redcross & \redcross & \greencheck \\
Sun \emph{et al.} (2025)~\cite{sun2025efficient} & \redcross & \redcross & \greencheck \\
Wan \emph{et al.} (2025)~\cite{WAN2025436} & \redcross & \redcross & \redcross \\
Yang \emph{et al.} (2025)~\cite{10.1145/3769748.3773346} & \greencheck & \redcross & \greencheck \\
\midrule
\rowcolor{ourrow}
\textbf{Ours} & \greencheck & \greencheck & \greencheck \\
\bottomrule
\end{tabular}
\end{table*}

Despite its central role, the systematic study of prompt engineering remains a fragmented area of research. While numerous surveys have reviewed the architecture and applications of SAM, a dedicated and systematic examination of prompt engineering—the core enabling mechanism behind SAM’s generalization capability—remains absent. As summarized in Tab. \ref{tab:sam_survey_comparison}, existing work largely focuses on SAM’s design and downstream performance across domains. For instance, several surveys \cite{zhang2023comprehensivesurveysegmentmodel, zhang2024surveysegmentmodelsam, jiaxing2025sam2imagevideosegmentation} present general overviews of SAM as a foundation model for segmentation. Other surveys \cite{10386032, 10386032} narrow their scope to SAM’s applications in medical image segmentation, while WAN \emph{et al.}~\cite{WAN2025436} targets remote sensing, and Sun \emph{et al.}~\cite{sun2025efficient} investigates compression and deployment optimizations. Although Yang \emph{et al.}~\cite{10.1145/3769748.3773346} offer a focused survey on point prompt engineering, their work is limited to a single prompt type and lacks a unified taxonomy across diverse strategies. This underscores the need for a comprehensive, prompt-centric perspective that systematically examines the role of prompt engineering in shaping SAM’s versatility and generalization.

To fill this gap, we conduct a systematic survey centered on prompt engineering within SAM and its growing ecosystem. Our review explores how prompt engineering has evolved beyond the original SAM formulation to support diverse application demands, including domain-specific adaptations (e.g., MedSAM \cite{gaillochet2024automating}, SurgicalSAM \cite{yue2024surgicalsam}), multimodal extensions (e.g., VL-SAM \cite{NEURIPS2024_80f48ffa}, ClipSAM \cite{LI2025129122}), and efficiency-oriented designs (e.g., Class-prompt Tiny-VIT \cite{DBLP:conf/cvpr/GuanDZ24}). Building on these observations, our key contributions are as follows:
\begin{itemize}
    \item We propose a hierarchical taxonomy of prompt engineering approaches for SAM (Fig.~\ref{fig:full_taxonomy}), systematically categorizing methods into geometric prompts, textual semantic prompts, and multimodal fusion prompts.
    \item We further analyze advanced automated generation strategies, including detector-based, reinforcement learning, and prototype learning techniques, to reveal how prompt engineering has evolved from manual annotation to data-driven adaptation across diverse domains.
    \item We identify promising future research directions, such as causal prompt engineering, collaborative multi-agent prompting, and diffusion-based progressive refinement, to stimulate further advancements in this evolving field.
\end{itemize}

To support our analysis and ensure reproducibility, we provide a clear definition of the literature scope and retrieval criteria. Our analysis covers works first made publicly available from SAM's official release in April 2023 through December 2025. Relevant literature was retrieved from major academic databases, including Google Scholar, Semantic Scholar, arXiv, IEEE Xplore, and the ACM Digital Library, using keywords such as ``Segment Anything'', ``promptable segmentation'', and ``prompt engineering''. We primarily include methods that leverage prompts to steer and customize the segmentation outcomes, while excluding task-specific models that lack prompt-driven interaction.

\begin{figure}[t]
\centering{
\begin{tikzpicture}[
    node distance=6mm and 8mm,
    every node/.style={rectangle, rounded corners, draw, align=center, font=\scriptsize},
    level1/.style={fill=blue!10, text width=28mm, minimum height=6mm},
    level2/.style={fill=green!15, text width=35mm, minimum height=8mm},
    level3/.style={fill=yellow!20, text width=62mm, minimum height=10mm},
    connection/.style={draw, -latex, thick}
]

\node[level1, text width=45mm, fill=red!10, rotate=90] (root) {Prompt Types};

\node[level1, above right=20mm and 11.5mm of root] (l1a) {Geometric Prompts};
\node[level1, below right=26mm and 5.5mm of root] (l1b) { Textual Semantic Prompts};
\node[level1, below right=50mm and 5.5mm of root] (l1c) { Multimodal Fusion Prompts};

\node[level2, above right=19mm and 5mm of l1a] (l2a1) {Manual Input and Heuristic Enhancement};
\node[level2, above right=3mm and 5mm of l1a] (l2a2) {Detector-Based Automatic Generation};
\node[level2, below right=2mm and 5mm of l1a] (l2a3) {Self-Prompting from Image Features};
\node[level2, below right=23mm and 5mm of l1a] (l2a4) {Optimization-Driven Generation Strategies};

\node[level2, above right=-2.5mm and 5mm of l1b] (l2b1) {Semantic Granularity-Based Prompt Construction};
\node[level2, below right=-0.5mm and 5mm of l1b] (l2b2) {Text-to-Geometric Prompt Generation};

\node[level2, right=5mm of l1c] (l2c1) {Joint Multimodal Feature Interaction};

\node[level3, right=3mm of l2a1] (l3a1) {
    PointPrompt \cite{quesada2024benchmarking}, 
    SAMAUG \cite{dai2024samaugpointpromptaugmentation} 
    EP-SAM \cite{song2024epsam}, 
    Few-shot-self-prompt-SAM \cite{wu2023self}, 
    CPC-SAM \cite{miao2024cross},
    RoBox-SAM \cite{huang2024robust},
    
    Automatic Segmentation Annotation \cite{10782995}

};

\node[level3, right=3mm of l2a2] (l3a2) {
    AM-SAM \cite{li2026sam}, 
    CrackESS \cite{wang2025crackess}, 
    Curriculum Prompting \cite{zheng2024curriculum}, 
    YOLO-SAM 2 \cite{mansoori2025self}, 
    SAM-SPB \cite{10648107},
    Char-SAM \cite{10888943},
    Diab~\emph{et al.} \cite{diab2025optimizing},
    RSPS-SAM \cite{liu2024rsps}
};

\node[level3, right=3mm of l2a3] (l3a3) {
    Automatic MedSAM \cite{gaillochet2024automating}, 
    De-LightSAM \cite{xu2025delightsammodalitydecoupledlightweightsam}, 
    AutoProSAM \cite{10943548}, 
    GroupPrompter \cite{10265054}, 
    SAM2Rad \cite{wahd2025sam2rad}, 
    TAVP \cite{yang2024tavptaskadaptivevisualprompt}, 
    APSeg \cite{he2024apseg},
    LA \cite{labelanything2025},
    SurgicalSAM \cite{yue2024surgicalsam},
    Chen~\emph{et al.} \cite{chen2024segmentation},
    APL-SAM \cite{shen2024adaptivepromptlearningsam},
    SAMIC \cite{nagendra2024samicsegmentincontextspatial},
    CycleSAM \cite{murali2024cyclesamoneshotsurgicalscene},
    SAID \cite{huang2025said},
    UCAD \cite{liu2024unsupervised},
    SPT-SAM \cite{10.1609/aaai.v39i12.33420},
    SAM-CD \cite{altam2023samcd},
    PUCD \cite{oh2023prototype},
    DED-SAM \cite{qiu2024ded}
};

\node[level3, right=3mm of l2a4] (l3a4) {
    TEPO \cite{10385291}, 
    AIES \cite{Hua_Optimizing_MICCAI2024},
    SAMRefiner \cite{lin2025samrefiner},
    GPRN \cite{hu2024relax},
    BioSAM \cite{10705688},
    GBMSeg \cite{liu2024feature},
    UR-SAM \cite{Zhang_2025_ICCV},
    UV-SAM \cite{zhang2024uv},
    EdgeSAM \cite{Zhou2025},
    SAM-MPA \cite{xu2024sammpa},
    MedSAM-U \cite{11201277},
    COMPrompter \cite{zhang2025comprompter}
};

\node[level3, right=3mm of l2b1] (l3b1) {
    AerOSeg \cite{dutta2025aeroseg},
    SP-SAM \cite{yue2024surgicalpartsamparttowholecollaborativeprompting},
    Talk2SAM \cite{vetoshkin2025talk2samtextguidedsemanticenhancement},
    TextSAM-EUS \cite{11375435}
    
};

\node[level3, right=3mm of l2b2] (l3b2) {
    CLISC \cite{ma2025clisc}, 
    GenSAM \cite{hu2024relax},
    PG-SAM \cite{zhong2025pgsampriorguidedsammedical},
    TP-DRSeg \cite{li2024tpdrseg},
    Class-prompt Tiny-VIT \cite{DBLP:conf/cvpr/GuanDZ24}
};

\node[level3, right=3mm of l2c1] (l3c1) {
    SEEM \cite{NEURIPS2023_3ef61f7e}, 
    ClipSAM \cite{LI2025129122}, 
    VLP-SAM \cite{jimaging12040143},
    VL-SAM \cite{NEURIPS2024_80f48ffa},
    TV-SAM \cite{jiang2024tv},
    FastSAM \cite{zhao2023fastsegment},
    SSPrompt \cite{huang2024learningpromptsegmentmodels},
    PPT \cite{dai2024curriculum},
    SegNext \cite{liu2024rethinking},
    SAA+ \cite{Cao_2025},
    EVF-SAM \cite{zhang2024evfsamearlyvisionlanguagefusion},
    LV-SAM \cite{Wu2025}
};

\begin{scope}[on background layer]

    \draw[connection] (root.center) -- ++(6mm,0) |- (l1a.west);
    \draw[connection] (root.center) -- ++(6mm,0) |- (l1b.west);
    \draw[connection] (root.center) -- ++(6mm,0) |- (l1c.west);

    \draw[connection] (l1a.east) -- ++(2mm,0) |- (l2a1.west);
    \draw[connection] (l1a.east) -- ++(2mm,0) |- (l2a2.west);
    \draw[connection] (l1a.east) -- ++(2mm,0) |- (l2a3.west);
    \draw[connection] (l1a.east) -- ++(2mm,0) |- (l2a4.west);
    
    \draw[connection] (l1b.east) -- ++(2mm,0) |- (l2b1.west); 
    \draw[connection] (l1b.east) -- ++(2mm,0) |- (l2b2.west);
        
    \draw[connection] (l1c.east) -- ++(2mm,0) |- (l2c1.west);

    \draw[connection] (l2a1.east) -- ++(2mm,0) |- (l3a1.west);
    \draw[connection] (l2a2.east) -- ++(2mm,0) |- (l3a2.west);
    \draw[connection] (l2a3.east) -- ++(2mm,0) |- (l3a3.west);
    \draw[connection] (l2a4.east) -- ++(2mm,0) |- (l3a4.west);
    
    \draw[connection] (l2b1.east) -- ++(2mm,0) |- (l3b1.west);
    \draw[connection] (l2b2.east) -- ++(2mm,0) |- (l3b2.west);
    
    \draw[connection] (l2c1.east) -- ++(2mm,0) |- (l3c1.west);

\end{scope}

\end{tikzpicture}}
\caption{Taxonomy of prompt types and generation methods.}
\label{fig:full_taxonomy}
\end{figure}

\section{Preliminaries}
To establish a common foundation for the prompt engineering techniques discussed in this survey, we first revisit the core architecture of the original SAM. While subsequent sections will explore specialized adaptations in various SAM variants, the prompt processing mechanism introduced in the original SAM serves as the fundamental framework underlying many later developments.

\subsection{Analysis of the SAM Architecture}
The Segment Anything Model (SAM) \cite{kirillov2023segment}, a promptable segmentation model, is designed to transfer zero-shot to new image distributions and tasks. As shown in Fig.~\ref{fig:sam-framework}, SAM is composed of 3 main components: an image encoder, a prompt encoder, and a lightweight mask decoder.

The image encoder processes the input image to generate an embedding that captures its essential features. This embedding serves as the foundation for the subsequent steps. The prompt encoder is responsible for embedding various types of prompts, such as points and boxes. These prompts direct the model to concentrate on pertinent regions of the image and identify the specific objects to be segmented. The lightweight mask decoder combines the image embedding and prompt embeddings to predict segmentation masks. This decoder employs a modified transformer decoder block followed by a dynamic mask prediction head. 

\subsection{Prompts in SAM}
\label{sec:prompt_in_sam}
\begin{figure}
    \centering
    \includegraphics[width=0.7\textwidth]{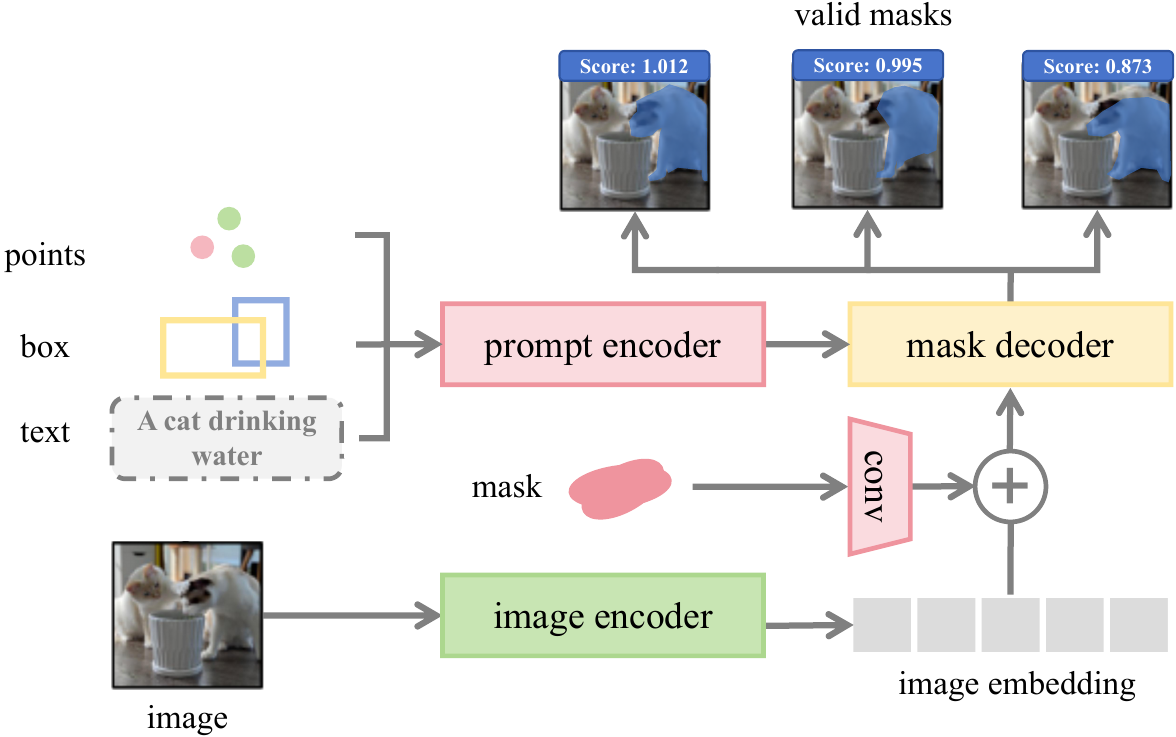}
    \caption{SAM framework  \cite{kirillov2023segment}.}
    \label{fig:sam-framework}
\end{figure}

SAM’s effectiveness in image segmentation relies on its prompt-guided mechanism. As illustrated in Fig.~\ref{fig:sam-framework}, it supports 3 fundamental interaction modalities: point-based, box-based, and mask-based prompts, each serving a distinct role in guiding the segmentation process.

For precise localization, point prompts allow users to mark specific image positions through inclusive points (identifying target regions) or exclusive points (excluding background areas). These prompts consist of coordinate-position pairs that undergo spatial encoding through a positional embedding layer. These encoded coordinates are then augmented with task-specific embeddings. The resulting representations, which are enriched with both spatial and semantic cues, are crucial for achieving accurate segmentation.
 
When segmentation requires spatial context rather than exact localization, box prompts provide an effective solution through axis-aligned bounding boxes parameterized by their opposing corner coordinates. SAM processes these boxes by converting corner points into positionally encoded vectors, which are then enhanced with additional embedding features to more robustly capture the geometry and structure of the enclosed region.

For complex shapes and multi-object scenarios, mask prompts offer the most detailed guidance. These can range from simple binary masks to sophisticated multi-class representations depending on segmentation needs. Accordingly, SAM processes these masks using convolution and embedding integration. The resulting features effectively delineate object boundaries and regions of interest \cite{kirillov2023segment}.

The flexibility of these prompt types enables SAM to adapt to various segmentation challenges, from simple object isolation to complex multi-instance identification tasks. This versatility makes it particularly valuable in applications requiring different levels of segmentation precision.

\begin{figure}
    \centering
    \includegraphics[width=0.96\textwidth]{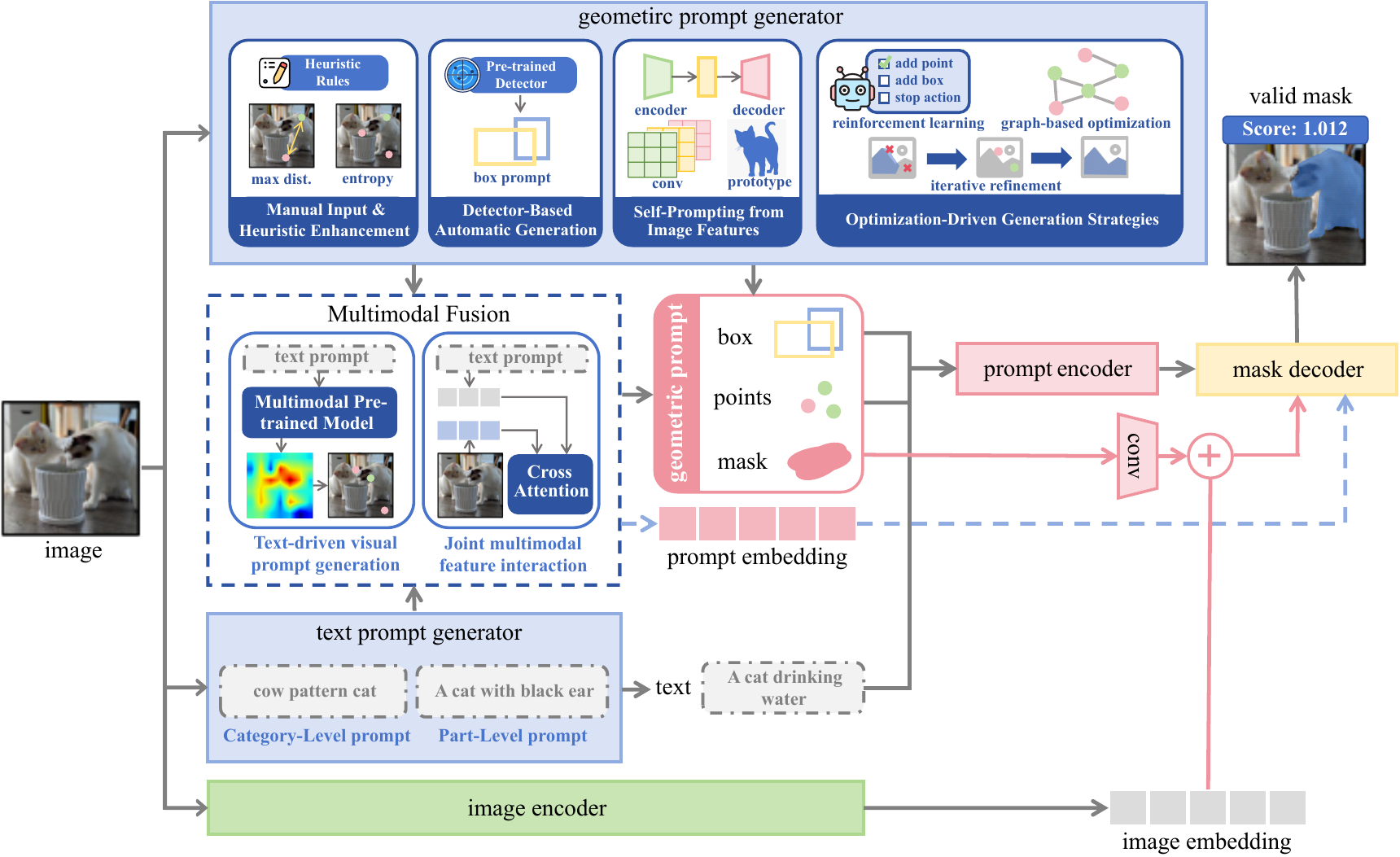}
    \caption{Overview of the SAM pipeline, illustrating key prompts and corresponding generation methods, comprising geometric prompts, textual semantic prompts, and multimodal fusion prompts.}
    \label{fig:overall-sam}
\end{figure}

\section{A Taxonomy of Prompt Engineering Methods}
To systematically organize SAM-based prompt engineering methods, our taxonomy is not merely based on the surface forms of prompts, but rather on their core functional roles and control mechanisms in the segmentation decision process. The classification logic in this survey is driven by a central question: how are the 2 key aspects of segmentation intent, namely what to segment and where to segment, expressed and conveyed through prompts? This perspective not only helps clarify the functional boundaries and evolutionary trends of various prompt types, but also provides a clear reference for future method design and task adaptation.
From this perspective, existing methods can be naturally categorized into 3 major types, ass illustrated in Fig.~\ref{fig:overall-sam}. 

Geometric prompts are dominated by spatial location and structural priors. They specify target regions directly via points, boxes, or masks, aiming to indicate ``where to segment.'' As such, they serve as the most fundamental and intuitive control interface for segmentation. 

Textual semantic prompts, in contrast, leverage high-level semantics and category knowledge to specify ``what to segment'' through textual descriptions. Within this taxonomy, the distinguishing criterion lies in the exclusivity of the user input source. Even if certain methods, such as text-to-geometric generation, involve cross-modal alignment internally to produce points or boxes, these signals serve strictly as spatial surrogates of the textual semantics rather than independent interaction modalities. Therefore, text remains the sole and primary driver of the segmentation intent.

Going further, multimodal fusion prompts incorporate geometric, visual, and linguistic signals simultaneously. Fundamentally different from text-driven approaches, multimodal prompts require signals from different dimensions to coexist as co-equal and independent external inputs, such as a user providing a manual click alongside a text query. This paradigm enables joint modeling within a unified representation space, allowing both what to segment and where to segment to be co-determined through multimodal interactions rather than being governed by a single signal source.

\subsection{Geometric Prompts: From Manual Annotation to Automated Generation}
\label{sec:geometric_prompts}
\subsubsection{Overview of Geometric Prompts}
While we have already briefly described the role of geometric prompts within SAM's interactive framework, this section focuses on their transformation into automatically generated inputs in SAM-based models. The perspective shifts from human-computer interaction to computational strategies for producing optimal points, boxes, and masks that reliably indicate ``where to segment''.

\begin{figure}[t]
    \centering
    \includegraphics[width=\linewidth]{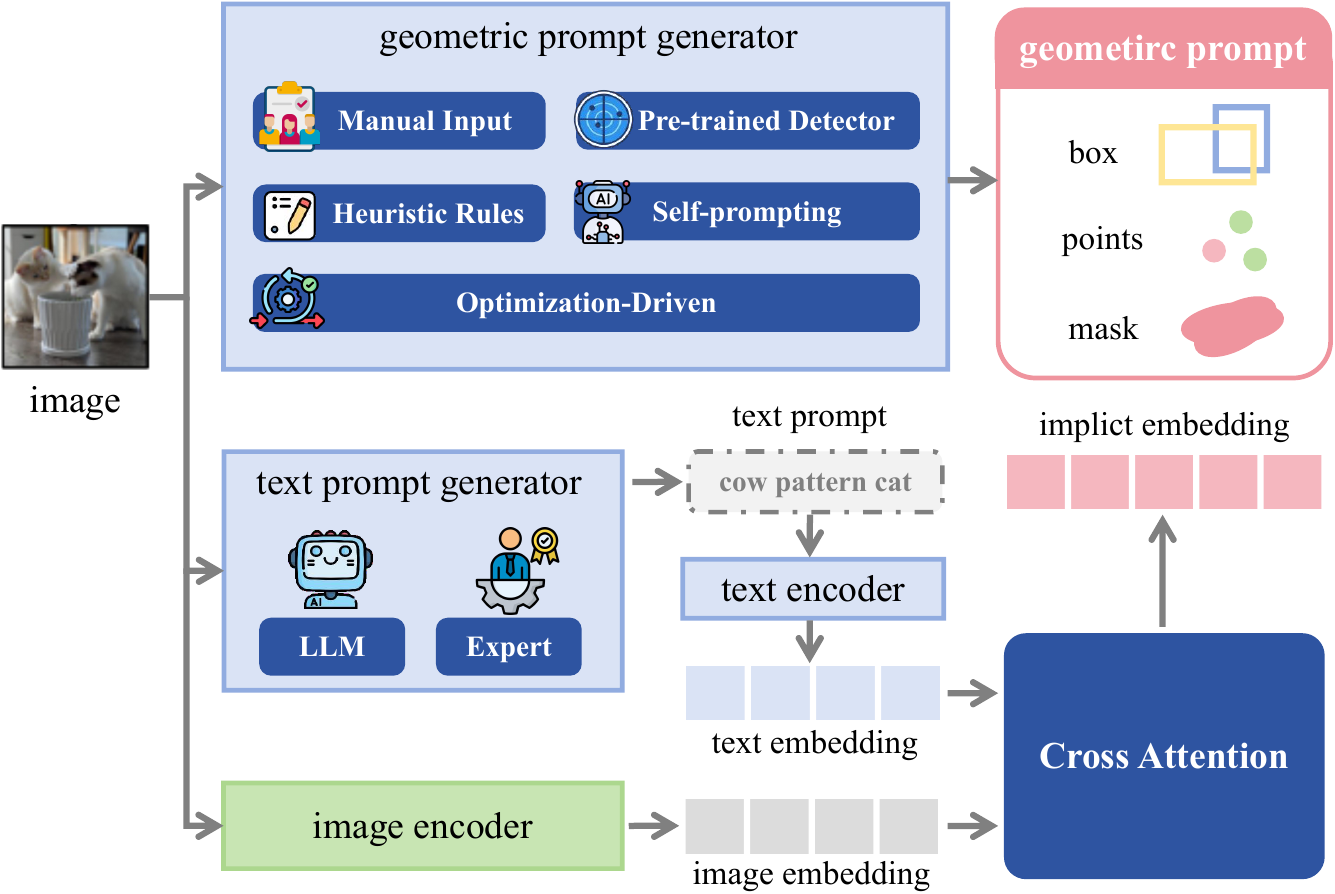}
    \caption{Structural workflow of geometric prompt generation strategies. The pipeline illustrates 4 mechanical routes: (1) manual input and heuristic enhancement leveraging geometric priors like maximum distance or entropy; (2) detector-based generation utilizing pre-trained models to provide bounding box anchors; (3) self-prompting mechanisms that map visual prototypes directly from image features; and (4) optimization-driven strategies employing reinforcement learning or graph-based refinement for iterative prompt calibration.}
    \label{fig:geometric_framework}
\end{figure}

Geometric prompts, the most intuitive and widely-used type of input, are often the first choice in domains such as medical imaging, remote sensing, and document analysis, where target regions can be readily specified via spatial locations. Consequently, the pursuit of automated prompt generation across these applications is driven by the need for greater precision, efficiency, and adaptability, aiming to reduce human intervention while maintaining or improving segmentation quality as illustrated in Fig. \ref{fig:geometric_framework}. Points, boxes, and masks each offer distinct trade-offs in terms of guidance granularity, annotation effort, and computational requirements.

\begin{itemize}
    \item Point prompts are the most flexible due to their low interaction cost, making them ideal for interactive segmentation and small targets. However, their effectiveness is highly sensitive to precise placement and can be limited for objects with ambiguous boundaries.
    \item Box prompts provide strong spatial constraints, making them highly effective for segmenting structured objects and naturally suited for detector-segmentation pipelines. By tightly constraining the search region, they guide SAM toward consistent foreground extraction. Their primary limitation, however, lies in heavy dependence on upstream detection accuracy and potential background noise for objects with irregular or diffuse boundaries.
    \item Mask prompts offer the richest spatial information, capable of guiding the model with detailed shape priors. They can serve as standalone prompts or as a basis for generating points and boxes. The challenge lies in the higher cost of obtaining a reliable initial mask.
\end{itemize}

\begin{figure}[t]
    \centering
    \includegraphics[width=\linewidth]{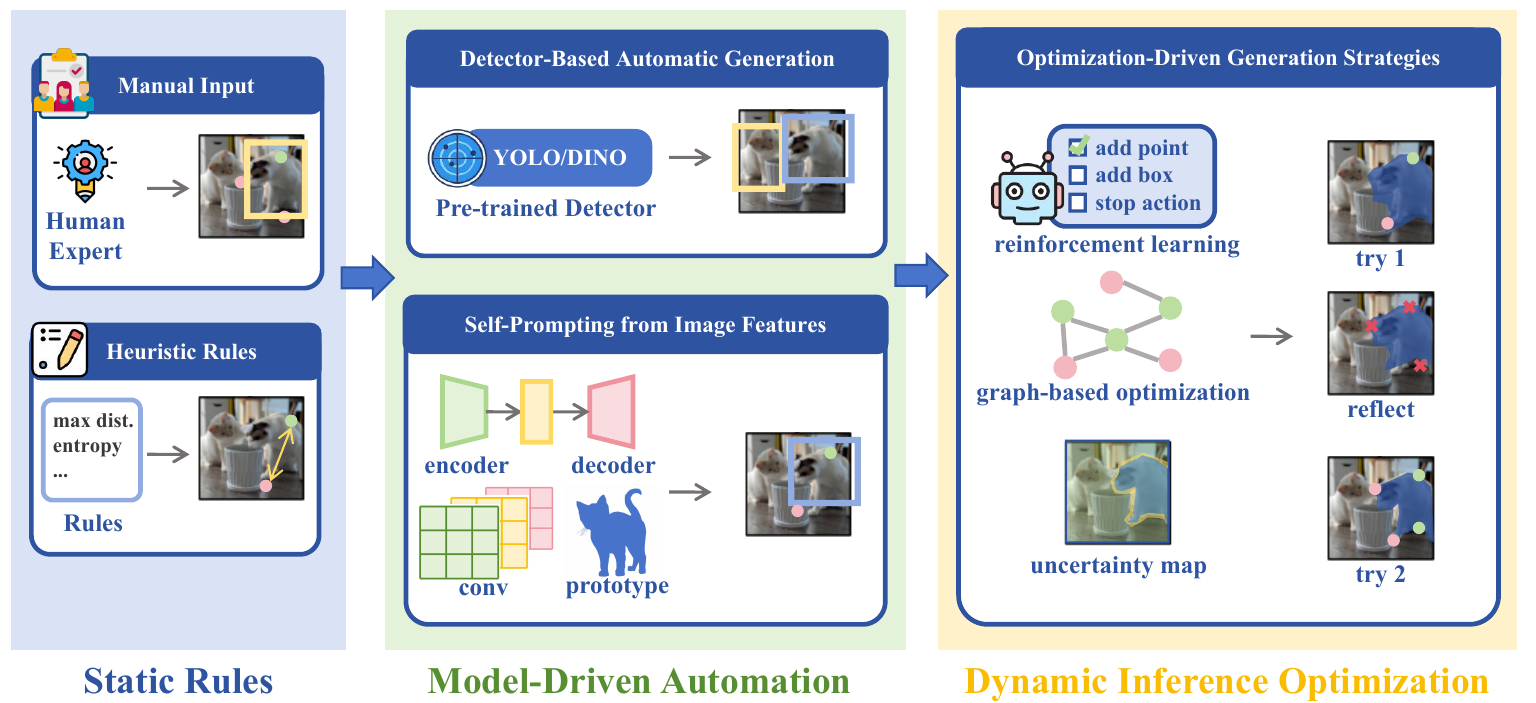}
    \caption{Logical evolution and interrelationships of geometric prompt generation paradigms. The flowchart illustrates the transition from human-centric or heuristic rules to detector-based and self-prompting automation, culminating in optimization-driven refinement. This progression represents a fundamental shift from static, single-pass generation to dynamic, multi-round iterative optimization for higher segmentation precision.}
    \label{fig:method_relation}
\end{figure}

Despite their utility, the engineering of geometric prompts faces several key challenges that motivate the automated methods discussed in this section. These challenges include prompt sensitivity, where minor deviations in prompt location can lead to large output variations; the difficulty of ensuring high-quality automatic generation from images or detectors; the need to maintain robustness across diverse domains; and the complexity of integrating multiple prompt types effectively without introducing conflicts. The following subsections explore the evolution of strategies designed to address these challenges, transitioning from manual annotation towards increasingly intelligent and automated generation paradigms.

The trajectory of geometric prompting has evolved along 2 primary axes: the nature of generation rules and the complexity of inference mechanisms. To systematically analyze this progression, we categorize existing automated strategies into 4 progressive paradigms, as illustrated in Fig.~\ref{fig:method_relation}. Manual and heuristic enhancement focuses on refining basic human inputs using static, predefined geometric logic, while detector-based generation seeks to replace manual effort with model-driven neural mappings from external object detectors. To further eliminate dependency on external models, self-prompting from features internalizes the generation process directly within the model's visual embedding space for higher autonomy. Ultimately, the field is advancing toward optimization-driven strategies, which introduce dynamic feedback loops and multi-round iterative refinement during the inference stage, elevating the process from simple automation to intelligent self-calibration. 

\subsubsection{Manual Input and Heuristic Enhancement}
\label{sec:maual_input_and_heuristic_enhancement}
Heuristic enhancement strategies build upon minimal manual prompts to improve segmentation robustness by algorithmically expanding or refining the initial inputs. These methods exploit inherent geometric properties, uncertainty measures, or preliminary intermediate model predictions to generate additional prompts or refine existing ones, thereby mitigating SAM’s sensitivity to single-point or single-box inputs in practical scenarios.

PointPrompt, a foundational framework for benchmarking these strategies, was established by Quesada \emph{et al.}~\cite{quesada2024benchmarking}. Their work provided crucial insights into human-AI collaboration for prompt generation by introducing a comprehensive benchmark that compares human-generated prompts with 6 automated heuristic strategies like K-Medoids and entropy sampling. A 29\% performance gap favoring human prompts in complex domains underscores the challenge of automating prompt generation and sets a benchmark for evaluating heuristic methods. The rich human-annotated dataset released by Quesada \emph{et al.} also serves as a valuable resource for subsequent research. Following this paradigm, numerous heuristic enhancement approaches have been developed.

A key approach focuses on point prompt augmentation and selection. SAMAUG \cite{dai2024samaugpointpromptaugmentation} generates multiple auxiliary points based on a coarse initial segmentation mask. It selects points using strategies such as random sampling within the mask, identifying regions with maximal entropy difference to resolve boundary ambiguity, choosing points farthest from the boundary to reinforce foreground centers, and highlighting salient regions via saliency maps. EP-SAM \cite{song2024epsam} employs pixel-level entropy maps to identify high-uncertainty areas, selecting points within these regions as positive prompts to enhance boundary recognition in weakly supervised settings.

Other works describe ways to derive prompts from preliminary model predictions and create a self-improving loop in which the model’s own predictions bootstrap further prompt generation. Few-shot-self-prompt-SAM \cite{wu2023self} generates a coarse mask using a simple classifier and identifies the pixel farthest from the boundary via a distance transform, which serves as a reliable center point prompt. CPC-SAM \cite{miao2024cross} selects both center points and random points from predictions on unlabeled data, using them as positive and negative prompts to enhance model stability through consistency regularization.

For box prompts, refinement techniques are used to correct inaccuracies. RoBox-SAM \cite{huang2024robust} predicts offset values to adjust the coordinates of a low-quality initial bounding box. By fusing image and prompt features via cross-attention, it effectively converts a poorly placed box into a high-quality one that tightly encloses the target object.

Beyond point and box prompts, some methods leverage structural information from masks to guide segmentation. Automatic Segmentation Annotation \cite{10782995} proposes iterative mask stitching and composite prompt engineering, integrating structural information from different masks to improve accuracy. By iteratively stitching masks and combining category attribution prompts, iterative mask stitching prompts, and heuristic point prompts, the mask quality is enhanced.

Overall, heuristic enhancement strategies bridge the gap between manual annotation and full automation, providing a robust and interpretable way to improve segmentation performance without requiring exhaustive manual labeling.

\subsubsection{Detector-Based Automatic Generation}
\label{sec:detector-bsaed_automatic_generation}
A highly effective approach to automating geometric prompt generation builds upon pre-trained object detectors. This paradigm, often forming a detect-then-segment pipeline, models such as YOLOv8 \cite{yolov8_ultralytics}, Grounding DINO \cite{liu2024grounding} and GLIP \cite{li2022grounded} are used to automatically produce spatial cues, primarily in the form of bounding boxes, which are subsequently provided as prompts to SAM. By leveraging the strong localization capabilities of these detectors, this strategy reduces the need for manual annotation while maintaining high-quality segmentation guidance.

The detector first identifies potential target objects within an image, and their corresponding bounding boxes are directly used as box prompts for SAM. For instance, AM-SAM \cite{li2026sam} and YOLO-SAM 2 \cite{mansoori2025self} directly employ YOLOv8 detection results as prompt inputs, providing accurate initial positional information for segmentation. Beyond simple replacement, AM-SAM further introduces a mask calibration module that performs element-wise multiplication between features to capture complex correlations, and then fuses the calibrated mask with the original output via weighted fusion to the final result. Similarly, CrackESS \cite{wang2025crackess} uses YOLOv8-generated bounding boxes as spatial prompts to guide the segmentation of cracks, overcoming the challenges posed by their complex morphologies. SAM-SPB \cite{10648107} integrates temporal consistency into optimization. It first generates textual descriptions of targets using Caption Anything \cite{wang2023caption}, and then applies Grounding DINO \cite{liu2024grounding} to produce candidate bounding boxes across frames. Subsequently, a memory mechanism selects the most reliable prompts to ensure stable segmentation over time.

Other works also describe ways to use the curricular or multi-stage prompting strategy to resolve potential conflicts between different prompt types and progressively refine coarse segmentations. Curriculum Prompting \cite{zheng2024curriculum} divides the process into 2 stages. In the coarse prompting stage, a detector such as Grounding DINO \cite{liu2024grounding} generates initial box prompts to produce a coarse mask. In the fine-grained prompting stage, keypoint detection networks, including HRNet \cite{sun2019deep} and ViTPose \cite{NEURIPS2022_fbb10d31}, generate point prompts targeting the boundaries of the coarse mask. Mask prompts are sometimes used as an intermediary to further facilitate boundary refinement. This phased approach enables progressive improvement of object localization and segmentation accuracy.

Char-SAM \cite{10888943} shows further specialization, tackling the precise segmentation of individual characters within text. It first uses a character-aware text detector for rough localization. Then, the word-level boxes are refined into precise character-level bounding boxes through bounding box separation, character class matching, and the watershed algorithm. These refined boxes are then used as visual prompts for SAM, and are further fused with glyph information to generate positive and negative point prompts, ensuring detailed character-level segmentation.

The primary advantage of detector-based methods is their high efficiency and scalability for processing large datasets. However, their performance is inherently contingent on the accuracy and generalization ability of the pre-trained detector. Errors in detection, such as missed objects or inaccurate bounding boxes, will directly propagate to and limit the quality of the final segmentation. Therefore, this approach is most effective in domains where reliable object detectors are readily available, mitigating this key dependency.

\subsubsection{Self-Prompting from Image Features}
\label{sec:self-prompting_from_image_features}
In contrast to detector-based methods that rely on external models, the self-prompt engineering learns to generate its own internal and adaptive prompt embeddings directly from image features. This approach aims to fully automate the prompting process by integrating it into an end-to-end learning framework, eliminating the need for pre-trained detectors and their associated annotation costs. It is particularly advantageous for domain-specific applications like medical imaging, where tailored prompt generation is crucial.

The core idea involves replacing SAM's original prompt encoder with a lightweight module that translates the image embedding into meaningful prompt representations. A common strategy is to generate both sparse and dense embeddings, mimicking the functions of point, box, and mask prompts. For instance, Automatic MedSAM \cite{gaillochet2024automating} proposes a lightweight prompt module that generates sparse embeddings through a fully connected layer to indicate key spatial positions, analogous to points, and dense embeddings to provide shape and scope information, analogous to low-quality masks. These generated embeddings are then fed directly into the mask decoder. Similarly, De-LightSAM \cite{xu2025delightsammodalitydecoupledlightweightsam} introduces a self-patch prompt generator that uses a convolutional network to downsample image embeddings into patch-level prompts and then upsamples them to form dense embeddings, all optimized via a binary cross-entropy loss to ensure the generated patches align with ground-truth annotations.

Beyond the basic generation of sparse and dense embeddings, researchers have developed more sophisticated architectures to learn prompt embeddings from image features, which incorporates deeper structural reasoning about the image content. For example, the AutoProSAM \cite{10943548} employs an encoder-decoder structure, which is similar to a 3D U-Net, to automatically learn prompt embeddings from the image encoder's feature maps. 

Other methods explicitly model the relationships within the feature space to generate more semantically coherent prompts. GroupPrompter \cite{10265054} proposes a 2-stage approach comprising ClusterPrompter and GroupPrompter. The ClusterPrompter module first performs hierarchical clustering on point features, dynamically aggregating them based on similarity to cluster centers to extract deep-level features. To address ClusterPrompter's limitations in capturing global context and cluster-center relationships, an attention mechanism is introduced to enhance global context. At each stage, learnable grouping tokens and image segments are processed by a lightweight transformer encoder. The updated segment tokens are then merged into new image segments based on embedding space similarity, effectively generating prompts by understanding the intrinsic structural groupings within the image. SAM2Rad \cite{wahd2025sam2rad} introduces a prompt predictor network to predict box coordinates, mask prompts, and high-dimensional embeddings from image features. The prompt predictor network employs a set of learnable tokens combined with positional encoding to capture spatial information. These tokens interact with the image features through a cross-attention mechanism, resulting in 2 outputs: a set of refined tokens and updated image features. The first 2 refined tokens are specifically decoded to predict bounding box coordinates, while the remaining tokens function as high-dimensional prompts. Simultaneously, the updated image features are used to generate an intermediate mask prompt. 

Another advanced direction within self-prompting leverages prototype learning to guide prompt generation. Instead of directly correlating image features, these methods learn a compact, representative prototype for each target category or structure. The segmentation prompt is then generated by comparing image features against these learned prototypes.

A prominent approach refines prototypes through sophisticated matching mechanisms. PGP-SAM \cite{yan2025pgp} refines intra- and inter-class prototypes to produce accurate prompts, integrating global semantic information into the process by its contextual feature modulation  module. Similarly, TAVP \cite{yang2024tavptaskadaptivevisualprompt} enhances foreground and background prototypes through multi-level feature fusion and bidirectional matching. APSeg \cite{he2024apseg} uses a dual prototype anchor transformation module and introduces a cycle-consistent selection mechanism to ensure the reliability and discriminative power of pseudo-prototypes. For multi-class scenarios, Label Anything \cite{labelanything2025} integrates mask, point, and box prompts, generates category prototypes through self-attention layers, category token pools, and bidirectional transformer, uses global average pooling to obtain class-example embeddings, and finally produces a single prototype via a class-example mixer.

This paradigm is also highly effective for specialized and domain-specific applications. SurgicalSAM \cite{yue2024surgicalsam} maintains a library of class prototypes. It computes a similarity matrix between the image embedding and these prototypes, using it as spatial attention to activate category-specific regions. In medical imaging, Chen~\emph{et al.} \cite{chen2024segmentation} uses image registration to align support masks and point prompts to the query image space, effectively using the aligned structures as anatomical prototypes. For more fine-grained prototype representation at the structural level in scanning probe microscope image segmentation, APL-SAM \cite{shen2024adaptivepromptlearningsam} employs detailed superpixel segmentation and unsupervised k-means clustering to encode local structures in the support image as compact visual prototypes, and generates prompt points based on centroid-level similarity.

An alternative yet related strategy focuses on leveraging deeper feature correlation and spatial alignment to infer prompts. SAMIC \cite{nagendra2024samicsegmentincontextspatial} converts user-provided prompts into saliency heatmaps and uses a hypercorrelation squeeze network \cite{min2021hypercorrelation} to capture semantic and geometric similarities between support and query images, converting the resulting heatmap into prompts via peak detection. CycleSAM \cite{murali2024cyclesamoneshotsurgicalscene} introduces cycle consistency constraints and multi-scale matching, verifying the correspondence between points and reference prototypes to filter erroneous prompts and generate high-confidence point prompts.

The primary strength of self-prompt engineering methods is their high degree of automation and domain adaptability without dependency on external models. However, their performance is heavily dependent on the quality of the learned feature representations and the design of the prompt generation module, which may affect their generalization to radically new domains outside their training data.

\subsubsection{Optimization-Driven Generation Strategies}
\label{sec:optimization-driven_generation_stratrgies}
Compared to heuristic strategies that rely on fixed rules, optimization-driven approaches aim to generate more effective prompts through either learning-based policies or numerical optimization techniques. Broadly, these methods can be divided into reinforcement learning (RL) paradigms and more general optimization-based paradigms, enabling greater flexibility.

Reinforcement learning–based approaches formulate prompt generation as a sequential decision-making problem, typically modeled as a Markov decision process. The central design components include the definition of the state space, the choice of available actions like adding points and boxes, and the construction of a reward function to guide policy learning.

TEPO \cite{10385291} defines the state space using medical image slices, the probability map from the previous timestep, and the history of interactive prompts. The action space consists of foreground points, background points, center points, and bounding boxes. The reward is computed as the difference in Dice scores between consecutive steps, thereby directly incentivizing segmentation improvement. AIES \cite{Hua_Optimizing_MICCAI2024} extends this idea by incorporating segmentation logits and interaction history into the state space. Its action space allows the agent to draw bounding boxes, click on mis-segmented regions, or terminate the interaction. The reward combines Dice improvement with penalty terms, encouraging the agent to achieve accurate results with fewer interactions. Wang \emph{et al.}~\cite{wang2025optimizing} further enhance the representation by integrating entropy maps, boundary gradient fields, and class distributions into the state. A deep Q-network is then used to approximate the Q-function, enabling the agent to continuously select regions as prompts in a human-like interaction loop. This iterative refinement process improves segmentation quality but also illustrates the high computational burden and the challenge of designing effective reward functions in RL-based methods.

Other optimization paradigms demonstrate a broader range of strategies tailored to specific application needs. In uncertainty-driven optimization, prediction uncertainties are modeled to guide prompt selection, ensuring that ambiguous or error-prone regions receive additional geometric cues for refinement. MedSAM-U \cite{11201277} incorporates an uncertainty-guided prompt adaptation module that predicts entropy maps, identifies ambiguous regions, and selects the top-ranked box and point prompts for refinement. This mechanism improves segmentation reliability by focusing on challenging boundary areas. UR-SAM \cite{Zhang_2025_ICCV} follows a similar philosophy, but it calculates prediction entropy directly and applies a class-specific confidence estimation strategy. This enables the system to filter out high-uncertainty regions and correct them with targeted prompts, thereby enhancing robustness.

In graph-based optimization, spatial and contextual dependencies among prompts are captured by modeling their relationships as a graph of structured entities, thereby enhancing semantic consistency and segmentation quality in a more principled manner. BioSAM \cite{10705688} leverages graph structures to impose spatial and contextual consistency. It begins with superpixel maps generated by a seeded watershed algorithm, then detects local peak points via Euclidean distance transformation to propose candidate prompts. These prompts are aggregated across adjacent superpixels using a graph neural network, which enhances spatial structure representation. BioSAM also incorporates structural priors such as intersection over union (IoU) scores and affinity graphs, allowing graph neural networks to establish contextually consistent prompt sets and thus strengthen semantic expressiveness. In contrast, GPRN \cite{hu2024relax} treats each visual prompt derived from initial masks as a graph node, forming a fully connected graph where node features are updated through a graph attention network. Spatial information is restored to obtain refined feature maps, which are then combined with support set masks to generate query predictions. During testing, GPRN employs an adaptive point selection module to filter high-confidence prompts from the initial predictions, feeding them back into the model for iterative refinement. This graph-driven optimization ensures semantic consistency among prompts and progressively improves segmentation quality.

In iterative refinement, new prompts are progressively sampled in mis-segmented regions, enabling continuous correction over multiple rounds. EdgeSAM \cite{Zhou2025} introduces a prompt-in-the-loop distillation mechanism. The model automatically samples new point prompts in mis-segmented regions for progressive multi-iteration correction. SAMRefiner \cite{lin2025samrefiner} addresses prompt inaccuracies through dual refinement mechanisms. For mask prompts, it introduces a Gaussian-style mask based on a distance transform, utilizing structural information such as the mask center point farthest from the background region to generate more accurate mask prompts. For box prompts, it proposes a context-aware elastic box that conditionally expands tight bounding boxes based on surrounding contextual information, avoiding quality degradation caused by false-negative pixels in coarse masks.

In matching and fusion optimization, prompt expressiveness is enhanced by aligning cross-modal or cross-hierarchical features and integrating complementary cues into more discriminative prompts. GBMSeg \cite{liu2024feature}, designed for glomerular basement membrane segmentation, enhances prompt discriminability through a multi-level optimization module that integrates forward and reverse matching, exclusive sampling, and hard negative sampling. It extracts block-level features using DINOv2 \cite{oquab2024dinov2}, constructs a distance matrix between blocks, and performs bidirectional matching to ensure precise alignment. SAM-MPA \cite{xu2024sammpa} improves prompt quality in few-shot scenarios by applying k-centroid clustering to select representative support images and masks, followed by an unsupervised B-spline elastic registration that estimates deformation fields to generate coarse query masks. These coarse masks then serve directly as mask prompts or be converted into box prompts through bounding box extraction, providing diverse geometric prompts with spatial alignment. UV-SAM \cite{zhang2024uv} employs SegFormer \cite{xie2021segformer}, a lightweight semantic segmentation model, to produce coarse segmentation masks of urban and rural areas, which are further processed to derive bounding boxes as prompts. COMPrompter \cite{zhang2025comprompter} fuses box prompts with boundary prompts by introducing boundary gradient extraction and a bidirectional guidance module. It further leverages high-frequency features extracted by a discrete wavelet transform, enriching the expressiveness of prompts and segmentation accuracy.

Overall, optimization-driven methods balance accuracy and efficiency by adaptively adjusting prompts based on task requirements. They represent an important step toward more intelligent and collaborative prompting systems. Nevertheless, they also face challenges such as high computational overhead, unstable convergence, and the difficulty of designing effective objective functions. Despite these obstacles, optimization-driven strategies highlight the future trajectory of prompt generation toward greater autonomy and adaptability.

\subsubsection{Summary and Comparative Analysis of Geometric Prompts} \label{sec:summary_geometric}
As summarized in Tab. \ref{tab:evolution_comparison}, these 4 paradigms exhibit fundamental differences in their operational logic, shifting from static, single-pass generation toward dynamic, closed-loop optimization.

\begin{table}[t]
\centering
\caption{Comparison of different geometric prompt generation paradigms across key technical dimensions, highlighting the evolution from static, human-centric rules to dynamic, autonomous inference mechanisms. }
\label{tab:evolution_comparison}
\resizebox{\linewidth}{!}{
\begin{tabular}{@{}lllll@{}}
\toprule
\textbf{Paradigm} & \textbf{Prompt Rule} & \textbf{Core Driver} & \textbf{At Inference} & \textbf{Interaction} \\ \midrule
Manual and Heuristic & Static & Human + Fixed Rules & Single-pass & High  \\
Detector-Based & Model-Driven & External Detectors & Single-pass & Low \\
Self-Prompting & Model-Driven & Internal Features & Single-pass & None \\
Optimization-Driven & Dynamic & Feedback Loop/RL & Multi-round & Autonomous \\ \bottomrule
\end{tabular}
}
\end{table}

Geometric prompting techniques have evolved considerably, shifting from initial reliance on manual fine-tuning, to incorporating heuristics for robustness, and finally advancing to autonomously generated prompts via object detection or end-to-end learning. This transformation is motivated not only by the imperative to reduce interaction costs but also by the need to mitigate SAM's inherent prompt sensitivity when processing specialized domain data.

Heuristic enhancement strategies, while maintaining high interpretability and requiring zero additional training costs, remain essentially reactive refinements. Their performance is strictly constrained by the quality of the initial manual seed prompt, making it difficult to recover from fundamental localization biases. In contrast, detector-based automatic generation considerably improves throughput for large-scale datasets by cascading mature object localization models. However, the robustness of such methods entirely depends on the generalization capabilities of the pre-trained detectors. In scenarios involving long-tail distributions or unseen professional contexts, failure of the detector can lead to a total collapse of the segmentation pipeline.

Self-prompting strategies successfully eliminate dependency on external detectors by encoding image features directly into prompt embeddings. This end-to-end adaptation is particularly effective in fields like medical imaging, where strong anatomical priors are required. Nevertheless, this deep coupling with domain-specific knowledge often comes at the expense of universal generalization, potentially resulting in inferior cross-scene transferability compared to the original SAM. 

Meanwhile, optimization-driven strategies, especially reinforcement learning-based approaches, represent the pinnacle of intelligent prompting by simulating the human observation-feedback-correction iterative loop. These methods exhibit superior robustness when handling highly challenging boundaries and complex topological structures. However, such performance gains are accompanied by a substantial increase in computational overhead; the complex search spaces and iterative mechanisms of reinforcement learning lead to high inference latency, which largely restricts their adoption in real-time applications.

\begin{figure}[t]
    \centering
    \includegraphics[width=0.9\linewidth]{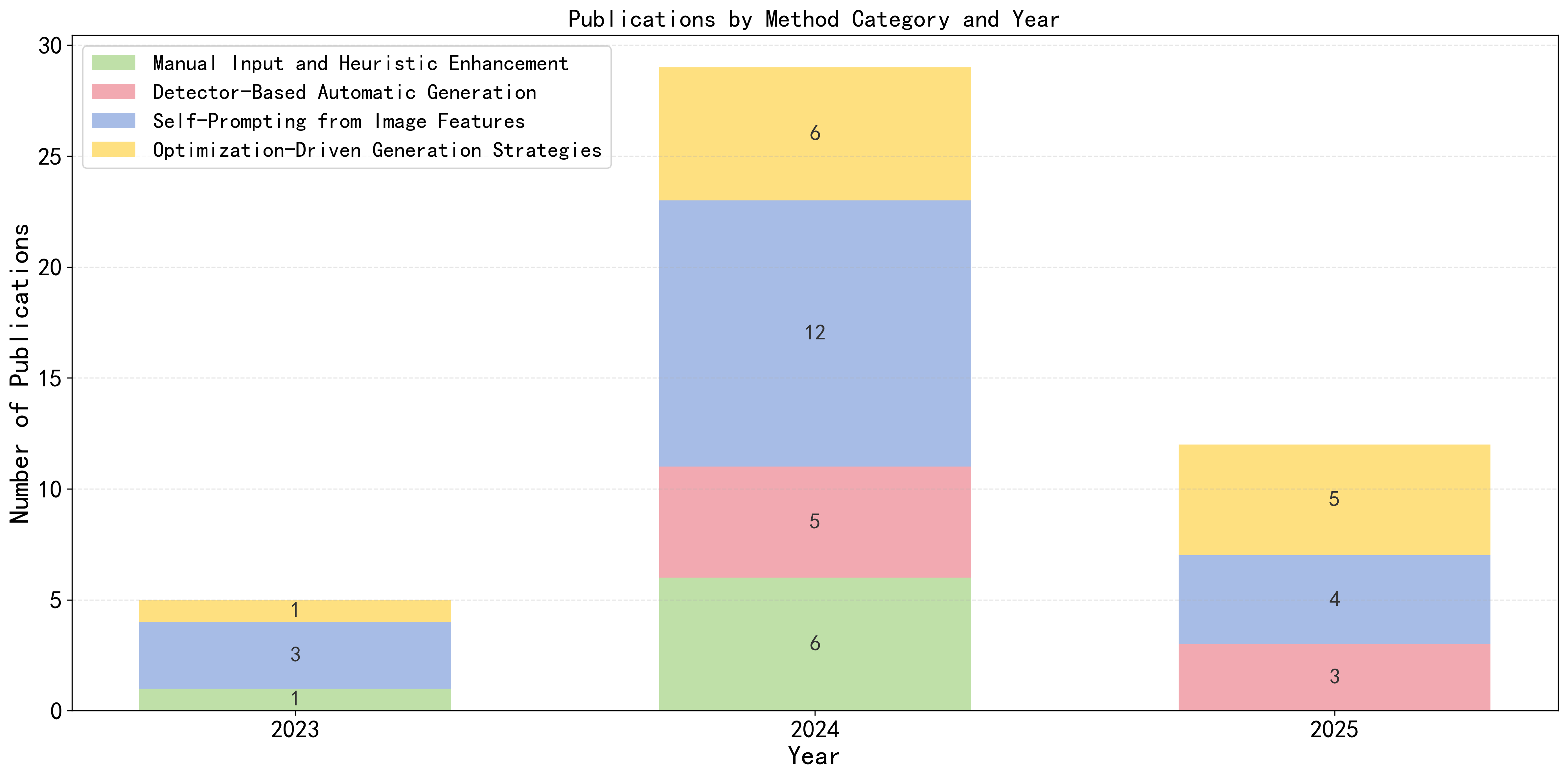}
    \caption{Evolutionary roadmap and quantitative distribution of geometric prompt engineering from 2023 to 2025 when this survey was conducted. This stacked bar chart shows an important paradigm shift: originating from heuristic enhancement of manual inputs, the research focus has rapidly transitioned toward endogenous self-prompting and high-precision optimization-driven refinement.}
    \label{fig:evolution_statistics}
\end{figure}

To further elucidate the evolutionary trajectory of geometric prompt generation strategies, we conducted a diachronic analysis of representative methods. As illustrated in Fig. \ref{fig:evolution_statistics}, the field underwent an important paradigm shift between 2023 and 2025. Originating from an initial phase (2023) primarily focused on heuristic enhancement, which aimed to refine manual inputs through simple geometric rules, the research landscape entered a period of rapid growth (2024) characterized by a surge in self-prompting techniques. This transition reflects a fundamental shift in research focus from relying on external interactions toward the endogenous generation of prompts within the model's feature space. In 2025, optimization-driven strategies became dominant. This trend indicates that the academic community is increasingly committed to enhancing segmentation precision through closed-loop refinement and iterative feedback, marking a leap in prompt engineering from preliminary automation toward high-level intelligence.

\subsection{Textual Semantic Prompts: Injecting High-Level Knowledge}
\label{sec:textual_prompt}
\subsubsection{Overview of Textual Semantic Prompts}
Textual semantic prompts provide SAM with high-level semantic guidance by positioning text as the primary driver of segmentation. Unlike textual cues in multimodal fusion, which serve mainly to enrich visual features alongside geometric hints, textual semantic prompts directly determine what should be segmented, either by injecting semantic knowledge into the feature space or by transforming textual descriptions into surrogate geometric prompts. Therefore, textual prompting does not imply the absence of geometric signals, but rather that such signals are derived from or subordinated to textual guidance within a more unified prompting framework.

\begin{figure}[t]
    \centering
    \includegraphics[width=\linewidth]{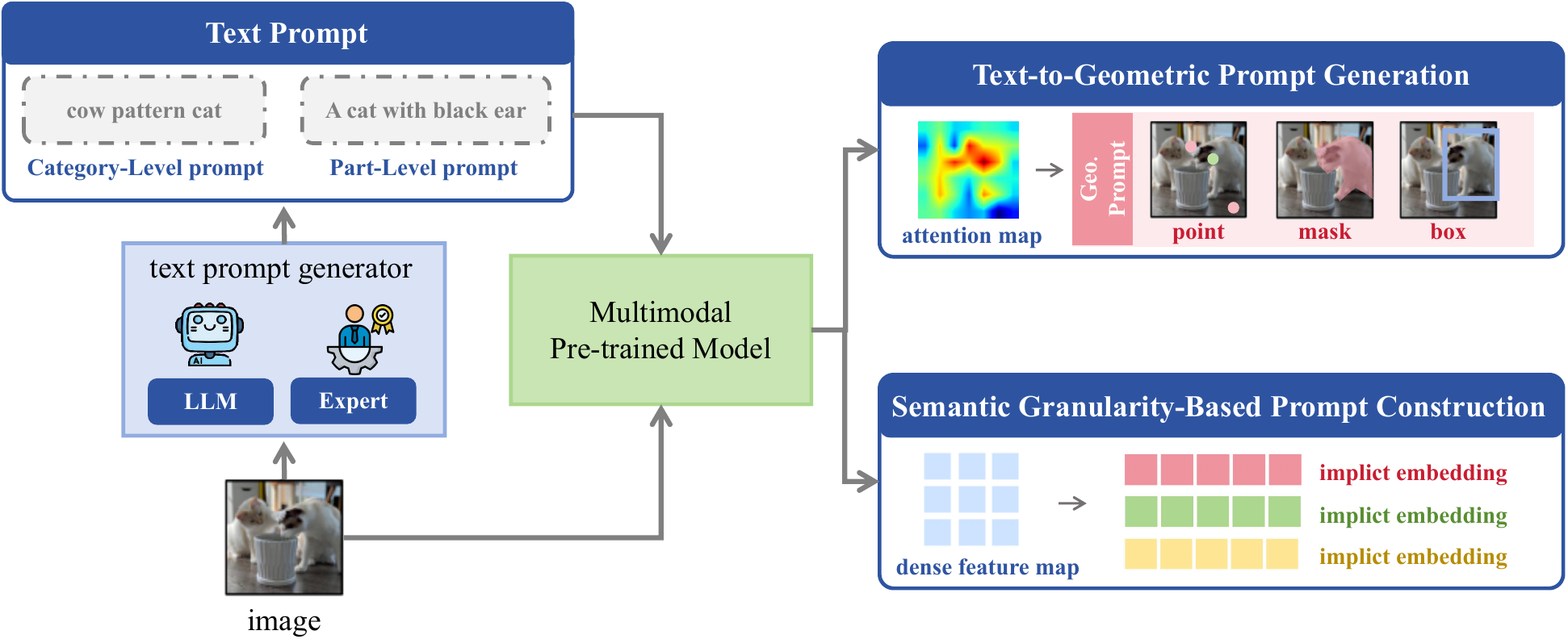}
    \caption{Architectural overview of textual semantic prompting. High-level knowledge is first formalized by LLMs or experts into category-level or part-level prompts. The framework then bifurcates into 2 inherent mechanisms: (1) semantic granularity-based prompt construction, which maps dense feature maps into implicit semantic embeddings for deep feature injection; and (2) text-to-geometric prompt generation, which utilizes cross-modal attention maps to sample spatial proxies (points, boxes) for SAM’s native interface.}
    \label{fig:textual_framework}
\end{figure}

While geometric prompts such as points and bounding boxes excel at providing spatial localization, they cannot distinguish between semantically similar objects or encode hierarchical part–whole relations. Textual semantic prompts fill this gap by encoding knowledge about object categories, structural components, and even domain-specific semantics, thus enabling SAM to go beyond purely spatial cues and support richer interpretive reasoning.

As shown in Fig. \ref{fig:textual_framework}, current research adopts 2 main strategies based on how semantic information is consumed by the model. Semantic granularity-based prompt construction focuses on internalized deep semantic injection, directly mapping text into dense embeddings or similarity maps within the feature space. In this paradigm, semantics function as implicit prior knowledge that reshapes the model's understanding of visual features, teaching the model ``what to segment''. In parallel, text-to-geometric prompt generation focuses on explicit surrogate coordinate transformation. It adopts a 2-stage strategy in which natural language is first aligned with visual features and then converted into explicit geometric prompts, such as points or boxes, that SAM natively consumes. In this case, the geometric signals act strictly as spatial surrogates for the linguistic intent, effectively letting the text perform the clicking on behalf of the user. These 2 routes represent different levels of semantic integration: the former prioritizes internal embedding alignment, while the latter primarily leverages the model's native geometric interface for downstream tasks.

Despite their advantages, textual semantic prompts also introduce new challenges. Textual descriptions may be ambiguous or coarse compared to pixel-level segmentation needs, and cross-modal alignment between text embeddings and visual features can be difficult, especially in specialized domains with rare or technical terminology.

\subsubsection{Semantic Granularity-Based Prompt Construction}
\label{sec:semantic_granularity-based_prompt_construction}
Semantic granularity-based prompt construction directly maps text semantics into the feature space of the segmentation model, generating dense embeddings or similarity maps as prompts. It typically leverages pre-trained vision-language models for cross-modal alignment, integrating textual guidance deeply within model's feature representations.

Category-level prompts guide segmentation using the overall category name of an object or region (e.g., ``bicycle'', ``tumor''), aiming to answer the fundamental question of ``what to segment''. The core advantages of this approach lie in its simplicity and generality, enabling efficient open-vocabulary segmentation. Talk2SAM \cite{vetoshkin2025talk2samtextguidedsemanticenhancement} transforms category names into dense semantic guidance. It employs a CLIP-based text encoder to convert category names into semantic embeddings, which are then projected into DINOv2's \cite{oquab2024dinov2} visual feature space through a learnable non-linear mapping. The model generates a dense semantic similarity map by comparing input image features with the projected text embeddings, highlighting regions relevant to the specified category. This similarity map is encoded by a trainable prompt encoder and injected into SAM-HQ's \cite{sam_hq} decoder, effectively integrating spatial structure with category semantics during segmentation. AerOSeg \cite{dutta2025aeroseg} uses 4 specialized prompt templates per category like ``A satellite image of a [CLS]'' that strongly couple the image modality with target categories. It generates text embeddings through CLIP encoding and employs multi-prompt fusion via average pooling to enhance robustness across diverse remote sensing scenarios and challenging environmental conditions.

For objects with complex structures, a single category-level prompt is often too coarse to guide the model towards fine-grained segmentation. Part-level prompts address this by decomposing the target object into its constituent parts and providing independent textual descriptions for each, thereby enabling the capture of detailed structures. In SP-SAM \cite{yue2024surgicalpartsamparttowholecollaborativeprompting}, surgical instruments are decomposed into constituent parts like ``shaft of large needle driver'' or ``tip of monopolar curved scissors'', and each part is represented as a structured textual prompt, enabling the model to translate abstract class names into concrete, visually discriminative cues. SP-SAM then employs a transfer MLP to map CLIP-encoded part embeddings into SAM's visual space, followed by spatial attention mechanisms to activate image regions corresponding to textual part descriptions. A dual attention fusion strategy handles compositional variations and occlusion, while hierarchical decoding enables simultaneous part-level and whole-instrument segmentation.

The pursuit of finer semantic granularity eventually encounters the practical constraints of data quality, particularly in domains like clinical medicine. The textual descriptions in reports are not only complex but also contain irrelevancies and ambiguities that can mislead a model. Therefore, robust semantic enhancement has emerged as a critical research direction, aiming to distill reliable segmentation signals from imperfect, real-world textual descriptions. TextSAM-EUS \cite{11375435} leverages a decomposition-recomposition strategy to addresses complex medical descriptions. It parses clinical reports into atomic semantic units and recomposes them into concise text pairs. A bidirectional cross-attention module enables interaction between different text prompt features, filtering redundant information while enhancing relevant semantic cues. The refined multimodal embeddings are directly fed into SAM's prompt encoder, improving robustness against noisy textual inputs.

\subsubsection{Text-to-Geometric Prompt Generation}
\label{sec:text-to-geometric_prompt_generation}
Text-driven visual prompt generation represents an important step toward reducing manual intervention by transforming natural language descriptions into spatially grounded visual cues. These approaches typically leverage multimodal vision-language models such as CLIP \cite{radford2021learning}, aligning text and image features in a shared embedding space to generate point or box prompts. Unlike purely manual geometric inputs, this strategy incorporates semantic context from textual descriptions, thereby enhancing interpretability and adaptability in complex scenarios.

A representative approach is to build on CLIP-based alignment pipelines. Models like CLISC \cite{ma2025clisc} employ domain-specific textual prompts like ``showing a tumor'' in medical scenarios and project text-image pairs into a shared space, producing pseudo-labels or coarse segmentation proposals. These are refined into visual prompts via techniques such as class activation mapping (CAM), adaptive mask data augmentation, and cross-modal decoding, which collectively enhance semantic grounding and guide SAM toward accurate localization. GenSAM \cite{hu2024relax} leverages BLIP2 \cite{zhu2024tmlr-chatgpt} to generate multi-perspective chain-of-thought keywords. Through an improved CLIP \cite{radford2021learning} self-attention mechanism, these keywords are aggregated into consensus heatmaps from which positive and negative point prompts are sampled, effectively grounding abstract language in spatial structures.

While CLIP-based alignment methods focus on aligning free-form text with image features, structured-knowledge driven approaches emphasize transforming domain-specific reports and expert priors into precise geometric or embedding-based prompts. PG-SAM \cite{zhong2025pgsampriorguidedsammedical} converts structured medical diagnostic reports into precise geometric prompts. Using CLIP for encoding and feature adaptation, its prompt generation module processes text embeddings alongside multi-sequence image features and statistical lesion distribution priors. The system outputs both bounding box prompts, which represent the minimum enclosing rectangles of lesion regions, and point prompts, which indicate core lesion locations. This provides strong spatial guidance derived from expert knowledge. TP-DRSeg \cite{li2024tpdrseg}, designed for diabetic retinopathy lesion  segmentation, implements a systematic text-to-prompt generation framework through its explicit text description strategy, converting medical prior knowledge into CLIP-understandable morphological descriptions. The 2-stage generation process first produces visual-language aligned prior maps via CLIP, then converts them into SAM-compatible dense and sparse embeddings through a class-specific prompt generator. The model further enhances this through prior-alignment injectors inserted into each SAM encoder block, enabling deep guidance of visual feature extraction by text prompts.

Other models focus on integrating categorical or learnable semantic priors into visual prompt generation. Class-prompt Tiny-VIT \cite{DBLP:conf/cvpr/GuanDZ24} uses a TinyVIT encoder \cite{wu2022tinyvit} to extract image features, predicts input image modality categories through a lightweight multi-layer perceptron network to generate category prompts, then fuses them with bounding box prompts before inputting to the mask decoder to produce per-category segmentation masks.

\subsubsection{Summary and Comparative Analysis of Textual Prompting Strategies} \label{sec:summary_textual_strategies}
Research on textual semantic prompting comprises 2 technical routes: deep semantic embedding and cross-modal coordinate transformation. The former focuses on maximizing the representational potential of textual information within the feature space, seeking to endow the model with profound interpretational capabilities through fine-grained semantic alignment. Conversely, the latter prioritizes the engineering implementation of automated workflows, striving for an optimal balance between model compatibility and systemic universality by transforming natural language into universal geometric anchors.

Strategies based on semantic granularity construction achieve end-to-end knowledge injection by mapping text directly into the feature space. Their core advantage lies in the ability to handle challenging part-whole hierarchical semantics, guiding SAM toward fine-grained segmentation by decomposing complex targets, such as specific functional components of surgical instruments. However, this deep feature alignment entails higher computational overhead, and its robustness is heavily contingent on the alignment precision of vision-language models within specialized domains. When confronted with clinical reports containing considerable noise or redundancy, these methods are susceptible to semantic drift.

In contrast, text-to-geometric prompt generation strategies emphasize systemic compatibility and interpretability. By converting natural language instructions into natively supported geometric anchors (e.g., points or boxes), these approaches can directly leverage SAM's pre-trained geometric reasoning without architectural modifications. While this surrogate-based generation process provides an intuitive view of intent localization, it faces a substantial risk of information loss, as rich textual nuances are often discarded when compressed into discrete spatial coordinates. Furthermore, these methods exhibit a pronounced error propagation effect, where minor deviations during the text-to-image alignment phase directly lead to misplaced geometric prompts, thereby limiting the final segmentation accuracy.

\subsection{Multimodal Fusion Prompts: Synergizing Vision and Language}
\label{sec:multimodal_fusion_prompts}
\subsubsection{Overview of Multimodal Fusion Prompts}
Multimodal fusion prompts establish a collaborative framework where textual semantics, geometric cues, and visual features serve as equal, complementary inputs. This paradigm moves beyond methods that prioritize one signal over another, instead integrating them into a unified representation. By allowing these modalities to interact on equal footing, the framework generates a more comprehensive and context-aware guidance signal.

The purpose of adopting multimodal prompts is to break through the inherent limitations of traditional prompting and to enhance SAM’s understanding and generalization in scenarios demanding complex semantics. In anomaly segmentation, combining text and visual prompts enables precise localization of abnormal regions while reducing linguistic ambiguity in a more controlled manner. In medical image segmentation, the fusion of modality-specific class prompts and spatial prompts improves segmentation accuracy across multimodal data by leveraging complementary information. For few-shot/zero-shot scenarios, multimodal prompts compensate for insufficient training data by incorporating textual semantics or reference image features, enabling effective segmentation of unseen categories. In real-time interactive tasks, lightweight prompt encoding and efficient feature fusion strategies considerably improve inference speed while maintaining accuracy.

This type of research addresses the challenge of modal alignment and feature fusion. Visual and textual signals reside in inherently different embedding spaces, and their integration requires a carefully designed mechanism such as cross-attention, prototype interaction, or shared encoding. These strategies aim to translate high-level semantic intent into precise pixel-level masks. 

\subsubsection{Joint Multimodal Feature Interaction}
\label{sec:joint_multimodal_feature_interaction}
Multimodal feature interaction aims to establish expressive joint representations between visual and textual embeddings, thereby achieving cross-modal alignment and complementarity across diverse scenarios. Unlike approaches that only generate prompts at the input level, these methods operate directly in the feature space, where cross-attention, self-attention, and prototype interaction modules enable context-aware and semantic–spatial fusion.

Fusion within a unified interaction space maps multimodal prompts into a shared semantic space, enabling deep, feature-level interaction through attention mechanisms. SEEM \cite{NEURIPS2023_3ef61f7e} unifies multimodal prompts in a single framework, integrating image features, text prompts, memory prompts, and visual prompts such as scribbles, boxes, and points. By dynamically sampling visual features and mapping text into a shared semantic space, SEEM employs cross-attention and self-attention mechanisms to retain historical segmentation information, enabling multi-round interaction and open-vocabulary segmentation. ClipSAM \cite{LI2025129122} employs a dual-path unified mutiscale cross-modal interaction  module to extract local row-column features and multi-scale global features from images. These features interact with textual features to generate coarse segmentation prompts, providing hierarchical semantic localization information for SAM. VLP-SAM \cite{jimaging12040143} utilizes vision-language models such as CLIP Surgery \cite{li2025closer} to embed reference images, target images, and text labels into a unified feature space. It then derives visual and semantic prototypes from these embeddings and integrates them via transformer-based interaction to achieve few-shot cross-modal fusion. Similarly, VL-SAM \cite{NEURIPS2024_80f48ffa} utilizes a vision-language model, CogVLM \cite{NEURIPS2024_dc06d4d2} to generate natural language descriptions of objects in images based on both image and text prompts. High-quality attention maps are derived by computing query-key similarity matrices during the description generation process. Positive and negative points are then sampled from these maps as point prompts. 

Sequential or hybrid multi-stage fusion decomposes the process into discrete stages, leveraging specialized models to progressively refine prompt generation. TV-SAM \cite{jiang2024tv} employs a 2-stage design. During text prompt generation, GPT-4's image and text encoders separately encode the input image and dialogue template, with features fused via cross-attention to produce text prompts. In the visual prompt generation stage, GLIP's image and text encoders \cite{li2022grounded} encode the input image and text prompts, respectively, and the fused features generate bounding box prompts, which are further refined via non-maximum suppression. FastSAM \cite{zhao2023fastsegment} first performs a full instance segmentation using YOLOv8-seg. It then selects target masks through 3 distinct strategies: using pixel-level matching for point prompts, relying on IoU matching for box prompts, and leveraging CLIP-based semantic similarity for text prompts. This multi-prompt approach enables the method to adapt to various interaction scenarios while maintaining high efficiency. PPT \cite{dai2024curriculum} employs a curriculum strategy. Through iterative interactions between learnable point queries and image or text features across multiple encoding stages, the model dynamically refines both the location and confidence of the generated prompts. SSPrompt \cite{huang2024learningpromptsegmentmodels} further addresses the limitations of pre-trained text encoders by introducing learnable embeddings for both spatial and semantic prompts, using weighted fusion to optimize alignment in high-dimensional spaces. 

Fusion enhanced by specialized strategies strengthens the core fusion process by introducing unique representations or domain-specific knowledge. SegNext \cite{liu2024rethinking} encodes visual prompts (clicks, boxes, scribbles) into 3-channel dense maps to preserve spatial details while supporting CLIP-generated language prompts, achieving fusion between dense maps and text embeddings through convolutional layers and self-attention modules. SAA+ \cite{Cao_2025} advances multimodal fusion by combining class-specific prompts defined by domain experts with anomaly saliency prompts that capture contextual deviations. These heterogeneous signals are aligned with image features in a shared embedding space using models like CLIP \cite{radford2021learning}, and further refined into point or box prompts through techniques such as CAM or adaptive mask data augmentation, thereby enabling robust and context-aware segmentation. EVF-SAM \cite{zhang2024evfsamearlyvisionlanguagefusion} uses BEIT-3 \cite{beit3} as an early-fusion multimodal encoder, jointly processing downsampled image patches and text tokens within each transformer layer to achieve cross-modal feature alignment. The fused embeddings are then projected to SAM’s prompt space and combined with spatial prompts, enabling efficient and semantically rich multimodal prompt generation. LV-SAM \cite{Wu2025}, developed for left ventricular ejection fraction estimation, fuses sparse prompts (points, boxes, text) and dense edge features through cross-attention and gating to balance contributions from different prompt types. 

In summary, joint multimodal feature interaction pushes prompt-based segmentation toward deeper and more coherent levels of semantic–spatial integration. While these approaches considerably boost segmentation accuracy and flexibility, they also introduce challenges such as high computational cost and complex training dynamics.

\subsubsection{Summary and Comparative Analysis of Multimodal Fusion Prompts} \label{sec:summary_multimodal}
Multimodal fusion prompts signify a shift from executing single-modality instructions toward achieving cross-modal semantic synergy. The core logic lies in dissolving the priority barriers between different modalities by constructing a unified interaction space, allowing visual, textual, and geometric signals to define targets through reciprocal and complementary means. This approach no longer relies on the precision of a single signal but leverages the redundancy of multi-source information to counteract the uncertainty of individual modalities in complex environments. Such a multi-dimensional collaborative strategy enables SAM to demonstrate semantic understanding depth far exceeding single-modality methods when handling open-world tasks, long-tail targets, and specialized scenarios with extremely low signal-to-noise ratios, such as industrial anomaly detection and complex surgical environments.

Under this paradigm, joint multimodal feature interaction represents the pursuit of feature-level deep alignment. By utilizing cross-attention mechanisms or shared encoders, these methods capture subtle correlations between textual semantics and spatial structures. This deep interaction allows the model to understand and execute highly abstract instructions, such as segmenting all regions with anomalous textures, and map them precisely to pixel coordinates, offering high flexibility and a high semantic upper bound. However, deep interaction introduces a considerable computational burden: since multimodal tensor operations are required within each feature layer, these models typically exhibit high inference latency. Furthermore, multimodal collaborative optimization increases the difficulty of model convergence and relies heavily on large-scale, high-quality alignment datasets, which are often difficult to acquire in professional domains.

In contrast, multi-stage and hybrid sequential fusion strategies adopt divide-and-conquer logic to achieve stronger engineering robustness and interpretability. By decoupling the workflow, they generate initial cues using specialized models such as GPT-4 or YOLO. These cues are subsequently refined through cascaded modules. Consequently, developers can independently select state-of-the-art pre-trained models for each stage. While this design simplifies training and provides a transparent error-tracing chain, it leads to cumulative errors and a sense of modal isolation. Since various modalities are processed independently in the early stages and lack real-time feature cross-talk, logical errors from the initial detectors and text parsers are difficult for subsequent modules to correct. Therefore, compared to joint interaction, these methods exhibit insufficient semantic integration depth when processing scenes where visual and textual information are highly overlapping and complex, potentially failing to fully realize the synergistic gains of multimodal prompting.

\subsection{Discussion and Synthesis: The Landscape of SAM Prompting}

\begin{table*}[t]
\centering
\scriptsize
\caption{Comprehensive comparative analysis of prompt engineering strategies for SAM. Methods are categorized by their primary prompt modalities, evaluating the trade-offs across Extra Training, Inference Latency, and Semantic Depth. The qualitative metrics follow a progressive scale: None $<$ Low $<$ Medium $<$ High $<$ Very High.}
\label{tab:comprehensive_prompt_comparison}
\renewcommand{\arraystretch}{1.4}
\begin{tabular}{lp{4.0cm}p{2.2cm}p{2.2cm}p{2.2cm}}
\hline
\textbf{Paradigm} & \textbf{Key Methods} & \textbf{Extra Training} & \textbf{Latency} & \textbf{Semantic Depth} \\
\hline
\rowcolor{ourrow}
\multicolumn{5}{c}{\textbf{Geometric Prompts: Spatial Localization Evolution}} \\
\hline
Heuristic & SAMAUG \cite{dai2024samaugpointpromptaugmentation}, RoBox-SAM \cite{huang2024robust} & None & Very Low & Low \\ \hline
Detector-based & YOLO-SAM \cite{mansoori2025self}, AM-SAM \cite{li2026sam} & Low & Low-Medium & Medium \\ \hline
Self-prompting & APSeg \cite{he2024apseg}, LA \cite{labelanything2025} & Medium & Medium & Medium-High \\ \hline
Optimization & TEPO \cite{10385291}, SAMRefiner \cite{lin2025samrefiner} & High & High & High \\
\hline
\rowcolor{ourrow}
\multicolumn{5}{c}{\textbf{Textual Semantic Prompts: High-Level Knowledge Injection}} \\
\hline
Deep Injection & Talk2SAM \cite{vetoshkin2025talk2samtextguidedsemanticenhancement}, SP-SAM \cite{yue2024surgicalpartsamparttowholecollaborativeprompting} & High & Medium-High & Very High \\ \hline
Geo. Surrogate & GenSAM \cite{hu2024relax}, CLISC \cite{ma2025clisc} & Medium & Medium & Medium \\
\hline
\rowcolor{ourrow}
\multicolumn{5}{c}{\textbf{Multimodal Fusion Prompts: Cross-modal Synergy}} \\
\hline
Joint Interaction & SEEM \cite{NEURIPS2023_3ef61f7e}, EVF-SAM \cite{zhang2024evfsamearlyvisionlanguagefusion} & Very High & Very High & Very High \\ \hline
Hybrid Seq. & TV-SAM \cite{jiang2024tv}, FastSAM \cite{zhao2023fastsegment} & Medium & Medium & High \\
\hline
\end{tabular}
\end{table*}

The taxonomy presented in this study delineates 3 types of SAM prompt engineering: geometric prompts, which focus on precise spatial localization; textual prompts, which emphasize high-level semantic injection; and multimodal prompts, which prioritize synergistic complementarity across different modalities. To delve deeper into the technical essence of these paradigms, we conduct a cross-dimensional comparative analysis as summarized in Tab. \ref{tab:comprehensive_prompt_comparison}.

First, in terms of information representation, the 3 paths exhibit a coexistence and transition between explicit coordinates and implicit embeddings. Geometric prompts are typical explicit representations; their advantage lies in extremely precise spatial guidance, serving as the cornerstone for handling structured objects. In contrast, textual prompts (especially deep-injection types) transform abstract semantics into implicit feature embeddings. This endows the model with the ability to understand content, addressing abstract instructions that geometric prompts cannot handle. Multimodal fusion further achieves the decoupling and recombination of both: leveraging text to determine object categories while utilizing geometry to refine boundary contours, thereby reaching an optimal balance between semantic depth and spatial precision.

Second, regarding robustness and error propagation mechanisms, different strategies exhibit distinct risk profiles. Detector-based and geometric surrogate strategies follow a sequential architecture, where performance is heavily contingent on the success of upstream tasks (detection or text alignment). Once a missed detection or semantic drift occurs in the front end, errors propagate irreversibly to the segmentation stage. Conversely, optimization-driven and joint interaction paradigms introduce feedback loops or multimodal redundancy. These allow the model to correct initial localization biases through iterative calibration or cross-modal complementarity during inference. While this closed-loop mechanism increases computational overhead, it considerably enhances survival capabilities in complex scenarios.

Finally, in the trade-off between universality and domain-specific customization, the paradigms follow a clear tiered distribution. Heuristic and surrogate-based strategies possess strong universality as they do not alter SAM's original architecture, making them the preferred choice for zero-shot tasks. However, in specialized domains such as medical imaging and remote sensing, this one-size-fits-all strategy often proves inadequate. Self-prompting and deep semantic injection sacrifice some degree of universality to learn domain-specific feature mappings, resulting in stronger adaptability for professional imagery. Consequently, prompt engineering is evolving from simple tool invocation into deep knowledge distillation.

In summary, while deep injection and joint interaction paradigms provide the highest semantic depth and robustness, they typically incur high training costs and inference latency. Future research should focus on how to effectively utilize the high-level reasoning capabilities of multimodal paths while maintaining the lightweight flexibility of early geometric prompting.

\section{Applications}
\subsection{Medical Image Analysis}
\subsubsection{Domain Challenges and Adaptation Logic of Prompting Strategies}
\label{sec:domain challenges-med}
To systematically address the multi-faceted challenges in medical imaging, we categorize existing prompting strategies into 3 core task-driven domains: anatomical strategies, which introduce spatial priors to ensure the structural consistency of normal tissues; pathological strategies, which leverage semantic enhancement to capture the heterogeneous features of abnormal lesions; and adaptive learning, which refines human-computer interaction workflows to achieve a seamless closed-loop from fundamental segmentation to clinical deployment, as summarized in Tab.~\ref{tab:sam_taxonomy_medical}.

\scriptsize
\renewcommand{\arraystretch}{1.4}
\begin{longtable}{>{\raggedright\arraybackslash}p{3.1cm}>{\raggedright\arraybackslash}p{2.1cm}>{\raggedright\arraybackslash}p{3.15cm}>{\raggedright\arraybackslash}p{3.15cm}>{\raggedright\arraybackslash}p{1.65cm}}
\caption{Domain-specific adaptation strategies: Addressing medical imaging challenges through prompt engineering.} \label{tab:sam_taxonomy_medical} \\
\hline
\textbf{Method} & \textbf{Pub.} & \textbf{Core Challenge} & \textbf{Mechanism} & \textbf{Modality} \\ \hline
\endfirsthead

\multicolumn{5}{c}{{\tablename\ \thetable{} -- continued from previous page}} \\
\hline
\textbf{Method} & \textbf{Pub.} & \textbf{Core Challenge} & \textbf{Mechanism} & \textbf{Modality} \\ \hline
\endhead

\hline \multicolumn{5}{r}{{Continued on next page}} \\
\endfoot

\hline
\endlastfoot

\rowcolor{ourrow}

\multicolumn{5}{c}{\textbf{Multi-scale Anatomical Segmentation}} \\

\hline

AutoProSAM \cite{10943548} & WACV 2025 & 3D spatial consistency & Adapter tuning & Box \\\hline

SAMCT \cite{lin2024samct} & IEEE TMI 2024 & Inter-slice discontinuity & Cross-slice attention & Box \\\hline

UR-SAM \cite{Zhang_2025_ICCV} & ICCV 2025 & Prompt-boundary ambiguity & Extremal point regression & Box \\\hline

LV-SAM \cite{Wu2025} & Med Biol Eng 2025 & Spatio-temporal inconsistency & Frame propagation & Multimodal \\\hline

Chen~\emph{et al.} \cite{chen2024segmentation} & WBIR 2024 & Low tissue-contrast & Morphological priors & Point, Box \\\hline

SAM2Rad \cite{wahd2025sam2rad} & CBM 2025 & Ultrasound speckle artifacts & Memory banking & Box, Mask \\\hline

All-in-SAM \cite{cui2024all} & JPCS 2024 & Massive tiny targets & Sparse sampling & Point \\\hline

SAC \cite{11218040} & IEEE TNNLS 2025 & Dense object adhesion & Active learning & Point, Box \\\hline

BioSAM \cite{10705688} & IEEE JBHI 2025 & Cross-modal domain gaps & Global adaptation & Box, Mask \\\hline

\rowcolor{ourrow}
\multicolumn{5}{c}{\textbf{Pathological and Lesion Detection}} \\
\hline
CLISC \cite{ma2025clisc} & ISBI 2025 & Diffuse boundaries and low contrast & Contrastive prototype & Point, Mask \\\hline

GBMSeg \cite{liu2024feature} & MICCAI 2024 & High tumor heterogeneity & Feature enhancement & Box \\\hline

Liu~\emph{et al.}~\cite{10656071} & IEEE NSS 2024 & Multi-modal inconsistency & Multi-source fusion & Box \\\hline

TextSAM-EUS \cite{11375435} & ICCV 2025 & Low EUS localization precision & Textual semantic learning & Text \\\hline

TP-DRSeg \cite{li2024tpdrseg} & MICCAI 2024 & Tiny lesion recall & 2-phase refinement & Point \\\hline

PP-SAM \cite{rahman2024pp} & CVPRW 2024 & Morphological diversity & Progressive selection & Point, Box \\\hline

YOLO-SAM 2 \cite{mansoori2025self} & ICASSP 2025 & Real-time tracking latency & YOLO-detector integration & Box \\\hline

\hline
\rowcolor{ourrow}
\multicolumn{5}{c}{\textbf{Clinically-driven Adaptive Learning}} \\
\hline

AIES \cite{Hua_Optimizing_MICCAI2024} & MICCAI 2024 & Suboptimal user click efficiency and interaction paths & Reinforcement learning agent & Point \\\hline

TEPO \cite{10385291} & IEEE BIBM 2023 & Error accumulation and noise in interactive prompts & Textual-enhanced optimization & Point \\\hline

Wu~\emph{et al.}~\cite{wu2023self} & DART 2023 & High cost of manual annotation and data scarcity & Self-prompting mechanism & Point, Box \\\hline

PGP-SAM \cite{yan2025pgp} & ISBI 2025 & Important domain shift in few-shot scenarios & Prototype-guided prompting & Point, Mask \\\hline

SAM-MPA \cite{xu2024sammpa} & NeurIPS 2024 (AIM-FM) &Few-shot medical segmentation and Limited supervision for multi-organ localization & Multi-scale prompt alignment & Box \\\hline

SAM-U \cite{deng2023sam} & MICCAI 2023 & Lack of feedback in iterative refinement & Uncertainty-aware feedback loop & Point, Mask \\\hline

MedSAM-U \cite{11201277} & IEEE TCSVT 2025 & Ambiguous boundaries in diverse medical scans & Uncertainty-guided masking & Box, Mask\\\hline

Automatic MedSAM \cite{gaillochet2024automating} & MedAGI 2024 & Heavy reliance on manual expert prompting & Automatic localization network & Box \\\hline

Swin-LiteMedSAM \cite{10.1007/978-3-031-81854-7_5} & CVPRW 2024 & High computational cost for clinical deployment & Hierarchical Swin Transformer & Box \\\hline

Guan~\emph{et al.}~\cite{DBLP:conf/cvpr/GuanDZ24} & CVPRW 2024 & Cross-modality feature alignment gap & Multi-modal latent prompting & Implicit Emb. \\\hline

TV-SAM \cite{jiang2024tv} & Big data min. anal. 2023 & Incomplete info from single-view imaging & Tri-view feature fusion & Point, Box \\\hline

SurgicalSAM \cite{yue2024surgicalsam} & AAAI 2024 & Complex instrument shapes and reflections & Prototype-based category prompt & Point, Mask \\\hline

CPC-SAM \cite{miao2024cross} & MICCAI 2024 & Large performance variance across users & Cross-prompt consistency check & Point, Box \\\hline

Curriculum Prompting \cite{zheng2024curriculum} & MICCAI 2024 & Difficulty in learning complex medical cases & Adaptive curriculum scheduler & Point, Box \\\hline

De-LightSAM \cite{xu2025delightsammodalitydecoupledlightweightsam} & IEEE TCSVT 2025 & Latency issues in real-time applications & Efficient spatial pruning & Box \\\hline

Wang~\emph{et al.}~\cite{wang2025optimizing} & Phys. Med. Biol. 2025 & Lack of standardized prompt protocols & Search-based optimization & Point \\\hline

RoBox-SAM \cite{huang2024robust} & MLMI 2024 & Model sensitivity to imprecise box inputs & Robust box-encoding & Box \\

\hline
\end{longtable}
\normalsize

\noindent \textbf{Multi-scale Anatomical Segmentation}\\
\noindent The target objectives of medical image segmentation span extreme spatial scales, ranging from macro-scale organs to micro-scale cellular tissues. The primary challenges in this field stem not only from discrepancies in physical target sizes but also from semantic loss and noise interference inherent in different imaging principles across scales. Although the vanilla Segment Anything Model (SAM) exhibits robust versatility on natural images, it encounters considerable dimensionality mismatch issues when handling multi-scale medical tasks: at the macro level, this manifests as a fracture in 3D spatial logic, while at the micro level, it results in sampling failure for high-density targets.

When processing 3D macro-imaging such as computed tomography (CT) and magnetic resonance imaging (MRI), the core pain point is SAM's 2D nature, which leads to inter-slice discontinuity. Anatomical organs are inherently continuous volumetric entities; however, 2D models treat them as isolated slices. This leads to jagged artifacts along the longitudinal axis (Z-axis) or anatomical logic conflicts in segmentation results like a kidney abruptly disappearing in a specific slice.

To compensate for this lack of 3D spatial semantics, SAMCT \cite{lin2024samct} introduces a lightweight task-indicator prompt encoder, using a cross-attention to transform feature information from adjacent slices into a dynamic prompt embedding. This approach endows the model with a sense of spatial depth, enabling 2D convolutional kernels to obtain cross-layer contextual constraints via prompt tokens. Addressing the clinical need for annotation efficiency, both AutoProSAM \cite{10943548} and UR-SAM \cite{Zhang_2025_ICCV} adopt automated localization strategies. Specifically, UR-SAM \cite{Zhang_2025_ICCV} emphasizes the concept of automated extreme point regression, mapping scans of different patients to a unified anatomical coordinate system through relative distance regression to automatically identify 3D extreme points and generate bounding boxes. These prompts, defined from a geometric perspective, directly solidify expert anatomical priors regarding organ locations into the model's localization operators.

In dynamic macro-imaging domains such as echocardiography, the challenge evolves into spatio-temporal inconsistency, where anatomical structures deform drastically over the cardiac cycle, accompanied by severe acoustic artifacts. LV-SAM \cite{Wu2025} ingeniously employs a temporal frame prompt propagation, using the segmentation mask of the previous frame as a dense prompt for the current frame. This logic utilizes the inertial prior of biological tissue motion to effectively suppress speckle noise common in ultrasound images. For bone-cartilage interfaces with extremely low contrast, Chen~\emph{et al.} \cite{chen2024segmentation} proposes a prompt filtering mechanism based on non-rigid registration, leveraging the relative stability of anatomical structures: even if image contrast is low, the relative positions of bones remain constant. By mapping expert prompts from a reference template to the target image via a deformation field, it guides SAM to lock onto boundaries in low-contrast environments.

At the micro-scale, such as in pathological sections, the core challenge shifts from spatial continuity to prompt overload and adhesion caused by a massive number of targets. Within a microscopic field of view, thousands of nuclei with extremely similar physical characteristics may exist simultaneously. Traditional manual point selection or simple grid sampling can lead to severe ambiguity in the model's semantic recognition. To address this issue, All-in-SAM \cite{cui2024all} proposes the logic of weak-prompt empowerment. In a physical sense, it treats extremely sparse point prompts as seeds, utilizing the powerful object-perception capabilities of pre-trained SAM to diffuse them into approximate annotations. The model is then fine-tuned for task-specific adaptation via a lightweight adapter. This method essentially uses SAM as a medium for knowledge distillation, solving the dilemma where large-scale high-quality annotations are unattainable at the micro-scale. In contrast, to handle extremely dense and blurry biological instances, BioSAM \cite{10705688} introduces a superpixel-guided prompt generation strategy. Rather than relying on isolated pixels, it utilizes a superpixel graph to capture the local clustering characteristics of biological tissues. By detecting local peaks via a Euclidean distance transform to precisely deploy prompt points, it ensures that each prompt locks onto an independent biological functional unit, circumventing misjudgments caused by dense sampling from a physical scale perspective.

Regarding the common instance adhesion challenge in microscopic images, where adjacent nuclei almost merge during imaging, SAC \cite{11218040} introduces an active learning-driven automated prompt generator. The underlying logic of this mechanism resides in uncertainty modeling: the model does not distribute computational power equally across all targets but automatically strengthens point prompt deployment in regions where prediction probability is most unstable. This targeted prompt engineering forces the model to focus on subtle texture fractures during training. Furthermore, the uncertainty rectification module proposed in UR-SAM \cite{Zhang_2025_ICCV} identifies high-entropy regions caused by tiny scales or blurry boundaries by adding random perturbations to initial prompts and calculating entropy maps. This method constructs a closed loop from prompt perturbation to result backtracking rectification, providing a rigorous reliability guarantee for the segmentation of anatomical structures with extreme scales and fragile boundaries.\\

\noindent \textbf{Pathological and Lesion Detection}\\
\noindent Unlike anatomical organs with relatively fixed morphology and locations, pathological lesions (e.g., tumors, polyps, ulcers) exhibit extreme heterogeneity. In clinical practice, lesions present highly irregular shapes and are often accompanied by extremely low imaging contrast, leading to blurred boundaries between pathological regions and surrounding healthy tissues. Due to a lack of pathological common sense, the vanilla SAM relies solely on geometric point or box prompts, which easily leads to over-segmentation or under-segmentation in such scenarios. Consequently, recent studies have constructed more robust lesion detection systems by introducing semantic priors, multi-modal information, and adaptive feature enhancement.

A core problem in lesion segmentation is lack of visual discriminative power. In endoscopic ultrasound (EUS) of the pancreas or early-stage diffusive tumor imaging, the grayscale intensity of lesions is extremely close to the background, making simple geometric prompts (points or boxes) ineffective. To address this challenge, TextSAM-EUS \cite{11375435} moves away from reliance on manual geometric prompts and introduces text semantic embeddings based on BiomedCLIP. Professional descriptors such as ``pancreatic tumor with irregular boundaries'' give the model pathological priors from a semantic dimension, enabling semantic-driven automated localization in noise-heavy ultrasound images. This text-to-image logic leverages the medical knowledge embedded in language models to compensate for the scarcity of visual features.

To address the segmentation overflow issue caused by blurred boundaries, the CLISC \cite{ma2025clisc} framework further proposes a contrastive prototype alignment strategy. It utilizes CLIP to generate image-level pseudo-labels and combines them with CAM to produce initial prompts. The core logic lies in strengthening the feature separation between tumor regions and normal tissues through contrastive learning. This constrains SAM's segmentation range via a semantic boundary in the feature space when physical boundaries are unclear, elevating lesion detection from simple pixel mining to the level of categorical discrimination.

Another considerable challenge for lesions is morphological uncertainty. Tumors of the same category can present vastly different infiltration patterns across different individuals. GBMSeg \cite{liu2024feature} targets the high heterogeneity of Glioblastoma by designing a feature-enhanced prompting mechanism. Rather than assuming lesions have regular geometric shapes, it captures the non-linear growth characteristics of tumors through multi-scale feature adapters, thereby dynamically adjusting the receptive field of the prompt tokens. When processing multi-modal MRI data, Liu~\emph{et al.}~\cite{10656071} emphasize the importance of multi-source prompt fusion. Since lesion appearances vary considerably across different sequences such as T1, T2, and FLAIR, this method generates composite box prompts by fusing multi-modal features, utilizing multi-dimensional information redundancy to offset the risk of lesions being inconspicuous in a single modality.

In lesion monitoring tasks within video sequences like colonoscopy, dynamic morphological changes and artifact interference are major obstacles. The YOLO-SAM 2 \cite{mansoori2025self} framework integrates a real-time object detector (YOLO) with the spatio-temporal memory network of SAM 2 to achieve automated polyp tracking. The logic involves using YOLO-generated bounding boxes as candidate prompts and leveraging SAM 2's memory bank to maintain feature consistency across frames. This strategy resolves the issue of prompt failure caused by drastic shape changes during polyp movement. Furthermore, targeting extremely tiny hemorrhagic spots in diabetic retinopathy, TP-DRSeg \cite{li2024tpdrseg} proposes a 2-stage refinement logic that injects medical prior knowledge via text prompts, excavating low-level features to capture these subtle lesions that are easily missed during diagnosis.\\

\noindent \textbf{Clinically-driven Adaptive Learning}\\
\noindent The research paradigm for SAM is shifting from manual intervention toward fully automated workflows and robust domain adaptation. Although SAM shows remarkable zero-shot generalization in natural scenes, its direct translation to clinical settings remains impeded by 3 critical challenges.

The most prominent hurdle is the domain discrepancy and the resulting limitations in feature representation. Medical imaging modalities, such as computed tomography (CT), magnetic resonance imaging (MRI), and ultrasound, possess distinct grayscale distributions and texture patterns that diverge considerably from natural images, making it difficult for the vanilla encoder to capture specific anatomical details. Medical images (e.g., ultrasounds, MRIs) are often accompanied by high-frequency noise, low contrast, and distributional variations across devices, making the original SAM highly susceptible to high-frequency perturbations and leading to unstable performance on cross-center datasets. Furthermore, the high-risk nature of clinical decision-making demands early-warning capabilities for misidentification. Existing methods typically address this by introducing lightweight hierarchical architectures to balance global context, combined with uncertainty estimation to construct a reliable visual feedback loop. Swin-LiteMedSAM \cite{10.1007/978-3-031-81854-7_5} replaces the vision transformer (ViT) with a hierarchical Swin Transformer, utilizing the shifted window mechanism to enhance feature extraction for complex anatomical structures. To tackle label scarcity, CPC-SAM introduces prompt consistency regularization, while SAM-MPA achieves unsupervised mask propagation across sequences through B-spline elastic registration. Additionally, MedSAM-U \cite{11201277} and SAM-U \cite{deng2023sam} propose multi-prompt triggered uncertainty estimation, generating uncertainty maps via Monte Carlo sampling to proactively flag high-risk regions where model confidence is low.

Beyond these distributional differences, the practical constraints of interaction cost and efficiency pose a considerable challenge. Clinical workflows demand rapid decision-making, yet an excessive reliance on manual bounding boxes or point prompts is not only labor-intensive but also introduces subjective variability due to the diverse annotation habits among clinicians. The performance of vanilla SAM is highly dependent on high-quality manual prompts, yet frequent manual interaction creates severe efficiency bottlenecks when processing large-scale clinical slices; moreover, due to a lack of domain knowledge, the model cannot understand complex diagnostic logic. Existing methods simulate physician interaction strategies through reinforcement learning or utilize multi-modal knowledge to achieve zero-shot automatic prompt generation. AIES \cite{Hua_Optimizing_MICCAI2024} and TEPO \cite{10385291} model the segmentation process as a Markov decision process, employing reinforcement learning agents to dynamically suggest optimal prompt forms based on the current state and autonomously decide when to terminate interaction. In terms of automation, TV-SAM integrates vision-language models to transform textual descriptions into visual guidance. SurgicalSAM introduces a prototype-based class prompt encoder, utilizing category features stored in a prototype library to replace manual indexing. Automatic MedSAM proposes a lightweight self-prompting module to derive dense prompts directly from image embeddings.

Compounding these efficiency issues is the intrinsic complexity of the targets themselves, particularly regarding blurred boundaries and minute scales. Pathological structures like early-stage tumors or cell nuclei often lack well-defined edges, causing traditional geometric prompts to be prone to false positives or edge shrinkage when contextual semantic information is absent. Existing methods optimize this by integrating morphological prior information to enhance image embeddings or by adopting multi-scale feature modulation to improve semantic alignment precision for tiny targets. RoBox-SAM \cite{huang2024robust} innovatively introduces a self-information extractor that enhances geometric constraints by fusing Canny contours, superpixels, and frequency-domain information, considerably correcting prompt biases at blurred edges. To address extremely small targets, PGP-SAM \cite{yan2025pgp} proposes a contextual feature modulation module that injects global semantics into multi-scale features to prevent the loss of critical lesions. Furthermore, the box-based scribble prompting strategy introduced by Swin-LiteMedSAM \cite{10.1007/978-3-031-81854-7_5} provides richer spatial priors than points or boxes, enhancing the capture precision for complex topological structures and micro-lesions.

\subsubsection{Quantitative Benchmarking and Computational Efficiency Analysis}
\normalsize
To provide a comprehensive overview of the current landscape, Tab. \ref{tab:sam_medical_performance} summarizes the quantitative benchmarking of various SAM-based adaptation strategies across diverse medical modalities. Unlike remote sensing or industrial defect detection, where pixel-level overlap (IoU/Dice) often suffices, medical imaging evaluation necessitates a multi-dimensional approach centered on clinical safety and biological fidelity. Since medical tasks are highly sensitive to edge overflow, standard volumetric overlap metrics are supplemented with boundary-aware indicators such as normalized surface distance (NSD) and the 95th percentile Hausdorff distance ($HD_{95}$). While a 5-pixel error might be negligible in general scenes, in a surgical context, it could signify the catastrophic misidentification of healthy nerves or critical vasculature. Furthermore, evaluation in this domain incorporates interaction efficiency (e.g., number of clicks) and uncertainty calibration (e.g., expected calibration error (ECE)) to account for the cognitive load on clinicians and the high-risk nature of diagnostic decision-making. The following sections analyze these performances through the lenses of anatomical fidelity, knowledge-driven generalization, and clinical utility.\\

\scriptsize
\renewcommand{\arraystretch}{1.4}
\begin{longtable}{>{\raggedright\arraybackslash}p{3.35cm} >{\raggedright\arraybackslash}p{3.9cm} >{\raggedright\arraybackslash}p{6.6cm}}
\caption{Quantitative benchmarking of SAM-based methods in medical imaging: Addressing domain-specific challenges via specialized adaptation logic.} 
\label{tab:sam_medical_performance} \\
\hline
\textbf{Method}& \textbf{Benchmark} & \textbf{Core Performance} \\ \hline
\endfirsthead

\multicolumn{3}{c}{{\tablename\ \thetable{} -- continued from previous page}} \\
\hline
\textbf{Method}& \textbf{Benchmark} & \textbf{Core Performance} \\ \hline
\endhead

\hline \multicolumn{3}{r}{{Continued on next page}} \\
\endfoot

\hline
\endlastfoot
\rowcolor{ourrow}
\multicolumn{3}{c}{\textbf{Multi-scale Anatomical Segmentation}} \\ \hline

AutoProSAM \cite{10943548} & BTCV, AMOS, CT-ORG, Institutional Pelvic& \makecell[l]{Dice: 87.15\%, NSD: 78.83\% (BTCV)\\
Dice: 88.65\%, NSD: 79.41\% (AMOS)\\
Dice: 85.12\%, NSD: 76.37\% (CT-ORG)\\
Dice: 91.30\%, NSD: 84.35\% (Institutional Pelvic)} \\\hline

SAMCT \cite{lin2024samct}& CT5M& \makecell[l]{Dice: 89.67\% (Avg), 86.32\% (unseen)}\\\hline

UR-SAM \cite{Zhang_2025_ICCV}& StructSeg, FLARE22 & \makecell[l]{Dice: 48.6\% (StructSeg, MedSAM)\\Dice: 58.6\% (FLARE22, MedSAM)} \\\hline

LV-SAM \cite{Wu2025}& CAMUS& Dice: 93.56\% (A2C), 94.20\% (A4C) \\\hline

Chen~\emph{et al.} \cite{chen2024segmentation}& Knee Joint MR & Dice: 89\% (Femur), 87\% (Tibia), 53\% (FemCart), 52\% (TibCart) \\\hline

SAM2Rad \cite{wahd2025sam2rad}& Musculoskeletal US & \makecell[l]{Dice: 78.2\%, IoU: 65.2\% (Hip, Hiera-Base+)\\
Dice: 63.5\%, IoU: 47.9\% (Wrist, Hiera-Base+)\\
Dice: 25.3\%, IoU: 25.1\% (Shoulder, Hiera-Base+)}\\\hline

All-in-SAM \cite{cui2024all}& MoNuSeg & Dice: 81.34\%, Best F1: 81.90\%, IoU: 68.10\% (completely labeled, 4\% training data)\\\hline

SAC \cite{11218040}& MoNuSeg, DSB& \makecell[l]{Dice: 84.03\%, IoU: 72.61\%, F1: 84.11\% (MoNuSeg) \\Dice: 93.04\%, IoU: 87.32\%, F1: 93.48\% (DSB)} \\\hline

BioSAM \cite{10705688}& AC3/AC4, CREMI, BBBC039V1, HEK293T & \makecell[l]{VOI: 0.7799, ARAND: 0.1797 (AC3/AC4)\\VOI: 0.5591, ARAND: 0.1188 (CREMI)\\VOI: 1.0490, ARAND: 0.1968 (BBBC039V1)\\VOI: 0.8826, ARAND: 0.1647 (HEK293T)}\\

\hline
\rowcolor{ourrow}
\multicolumn{3}{c}{\textbf{Pathological and Lesion Detection}} \\
\hline
CLISC \cite{ma2025clisc}& BraTS2020 & Dice: 85.60\%, $HD_{95}$: 6.72 mm \\\hline

GBMSeg \cite{liu2024feature}& TEM Glomerular Basement Membrane& Dice: 87.27\% (one-shot) \\\hline

Liu~\emph{et al.}~\cite{10656071}& BraTS2018 & Dice: 65.29\%, HD: 21.12 mm (T=10)\\\hline

TextSAM-EUS \cite{11375435}& Endoscopic Ultrasound Database of the Pancreas & Dice: 82.69\%, NSD: 85.28\% (automatic)\\\hline

TP-DRSeg \cite{li2024tpdrseg}& IDRiD, DDR & \makecell[l]{mDice: 49.72\%, AUC-ROC: 95.94\% (IDRiD)\\ mDice: 38.78\%, AUC-ROC: 94.12\% (DDR)} \\\hline

PP-SAM \cite{rahman2024pp}& Kvasir (ClinicDB, EndoScene, ColonDB) & \makecell[l]{DICE: 74.5\% (1-shot, 50px perturbation)\\
DICE: 77.0\% (5-shot, 50px perturbation)\\
DICE: 81.6\% (10-shot, 50px perturbation)} \\\hline

YOLO-SAM 2 \cite{mansoori2025self}& Kvasir-SEG, CVC-ClinicDB, CVC-ColonDB, ETIS, CVC-300, SUN-SEG, PolypGen & \makecell[l]{mDice: 95.1\%, mIoU 90.9\% (CVC-ClinicDB)\\
mDice: 86.6\%, mIoU 76.4\% (Kvasir-SEG)\\
mDice: 91.8\%, mIoU: 84.8\% (CVC-ColonDB)\\
mDice: 90.4\%, mIoU: 90.4\% (ETIS)\\
mDice: 88.9\%, mIoU: 81.7\% (CVC-300)} \\\hline
\rowcolor{ourrow}
\multicolumn{3}{c}{\textbf{Clinically-driven Adaptive Learning}} \\
\hline
AIES \cite{Hua_Optimizing_MICCAI2024}& Brats2020, Polyp, Spleen & \makecell[l]{Dice: 82.03\% (Brats2020, 6 steps)\\ Dice: 90.46\% (Polyp, 6 steps)\\ Dice: 89.22\% (Spleen, 6 steps)} \\ \hline

TEPO \cite{10385291} & BraTS2020 & \makecell[l]{Dice: 83.42\% (TEPO-7, 9 steps)\\Dice: 84.49\% ((TEPO-9, 9 steps)} \\ \hline

Wu~\emph{et al.}~\cite{wu2023self}&Kvasir-SEG, ISIC-2018&\makecell[l]{Dice: 62.78\%, IoU: 53.36\% (Kvasir-SEG)\\ Dice: 66.78\%, IoU: 55.32\% (ISIC-2018)}\\\hline

SAM-MPA \cite{xu2024sammpa}& Breast US, Chest X-ray& \makecell[l]{Dice: 74.53\% (Breast US, 10-shot)\\Dice: 94.36\% (Chest X-ray, 10-shot)} \\\hline

SAM-U \cite{deng2023sam}& REFUGE Challenge& \makecell[l]{Dice: 72.6\%, ECE: 1.2\% (ViT-L, $\alpha$=0.05)\\
Dice: 69.0\%, ECE: 1.5\% (ViT-L, $\alpha$=0.10)\\
Dice: 75.1\%, ECE: 1.3\% (ViT-L, high-quality)} \\\hline

MedSAM-U \cite{11201277}& ISIC-2017, CT2US, KiTS23, BraTS 2021, Kvasir-SEG  & Dice: 86.1\%, IoU: 78.4\%(3B overlap 0.75, Avg.) \\ \hline

PGP-SAM \cite{yan2025pgp}& Synapse multi-organ CT, Ventricle CT & \makecell[l]{Dice: 78.75\% (Synapse multi-organ CT)\\Dice: 76.39\% (Ventricle CT)} \\\hline

Automatic MedSAM \cite{gaillochet2024automating}& HC18, CAMUS, ACDC& Dice: 85.23\% (HC18), 88.38\% (CAMUS-LV), 73.56\% (CAMUS-LA), 58.96\% (ACDC-RV), 80.37\% (ACDC-LV) \\ \hline

Swin-LiteMedSAM \cite{10.1007/978-3-031-81854-7_5} & CVPR 2024 Challenge& \makecell[l]{Dice: 86.70\%, NSD: 88.55\% (Val.)\\Dice: 81.93\%, NSD: 84.61\% (Test)} \\\hline

Guan~\emph{et al.} \cite{DBLP:conf/cvpr/GuanDZ24}& CVPR 2024 Challenge & \makecell[l]{Dice (Avg): 86.89\%, NSD (Avg): 88.04\%} \\\hline

TV-SAM \cite{jiang2024tv} & Polyp benchmark, ISIC 2018, WBC dataset, BUSI dataset, TN3K dataset, COVID-19 Database (X-ray and CT), CHAOS dataset (T1 and T2 MRI) & Dice: 83.1\% (Polyp), 85.3\% (ISIC), 96.8\% (WBC), 75.1\% (BUSI), 58,8\% (TN3K), 78.8\% (X-ray), 63.6\% (CT), 47.0\% (T1 MRI), 48.8\% (T2 MRI)\\\hline

SurgicalSAM \cite{yue2024surgicalsam} & EndoVis2018, EndoVis2017& \makecell[l]{IoU: 80.33\%, mc IoU: 58.87\% (EndoVis2018) \\ IoU: 69.94\%, mc IoU: 67.03\% (EndoVis2017)}  \\ \hline

CPC-SAM \cite{miao2024cross} & BUSI, ACDC & \makecell[l]{
    Dice: 71.20\%, $HD_{95}$: 100.22 px (BUSI, 10 labels)\\
    Dice: 72.41\%, $HD_{95}$: 96.26 px (BUSI, 20 labels)\\
    Dice: 85.56\%, $HD_{95}$: 9.19 mm (ACDC, 1 patient)\\
    Dice: 87.95\%, $HD_{95}$: 5.80 mm (ACDC, 3 patients)
} \\ \hline

Curriculum Prompting \cite{zheng2024curriculum} & TN3K, Kvasir, QaTa-COV19 & \makecell[l]{mDice: 84.43\%, mIoU: 76.37\% (TN3K)\\ mDice: 93.67\%, mIoU: 89.44\% (Kvasir)\\ mDice: 79.83\%, mIoU: 70.27\% (QaTa-COV19)} \\\hline

De-LightSAM \cite{xu2025delightsammodalitydecoupledlightweightsam}& ISIC-2018, PCXA, DRIVE, CVC-ClinicDB, UDIAT, DSB-2018 & \makecell[l]{Dice: 88.52\%, HD: 92.42  (ISIC-2018)\\ Dice: 96.83\%, HD: 40.68 (PCXA)\\ Dice: 82.42\%, HD: 62.64 (DRIVE)\\ Dice: 92.93\%, HD: 56.32 (CVC-ClinicDB)\\ Dice: 85.28\%, HD: 82.32 (UDIAT)\\ Dice: 92.24\%, HD: 85.93 (DSB-2018)}\\\hline

Wang~\emph{et al.}~\cite{wang2025optimizing}& Ovarian, Renal, Lung, Breast Tumors& Dice: 59.5\% (Ovarian), 50.5\% (Renal), 54.0\% (Lung), 51.5\% (Breast) \\\hline

RoBox-SAM \cite{huang2024robust}& Multi-modality dataset (99,299 images, 5 modalities, 25 organs) & Dice: 90.09\% (MRI), 88.02\% (CT), 83.17\% (Ultrasound), 93.77\% (Colonoscopy), 91.72\% (Fundus) \\
\hline
\end{longtable}
\normalsize

\noindent \textbf{Performance Reliability and Anatomical Fidelity}\\
\noindent The fundamental challenge of medical segmentation lies in overcoming the lack of anatomical logic caused by imaging artifacts and biological variations. Although the vanilla SAM possesses strong geometric perception in natural images, it often leads to fractures or overflows in anatomical structures when facing medical volumetric data due to a lack of 3D spatial constraints. Addressing the issue of 3D spatial consistency, SAMCT \cite{lin2024samct} achieved an average Dice score of 89.67\% on the large-scale CT5M dataset covering 118 anatomical categories and maintained a performance of 86.32\% on the unseen BTCV dataset. Notably, the improvement of SAMCT was particularly large when handling extremely difficult-to-segment targets such as the gallbladder and esophagus, proving that the cross-slice attention mechanism can resolve anatomical ambiguities caused by tissue overlap through global context. Additionally, regarding adaptation to extreme scales, UR-SAM \cite{Zhang_2025_ICCV} achieved a Dice improvement of 10.7\% to 13.8\% on small organs such as the spinal cord and adrenal glands, reinforcing the value of its uncertainty rectification strategy in preventing minute structures from being overlooked. The effectiveness of physical priors was also confirmed in Chen~\emph{et al.}; the model achieved a Dice greater than 87\% for high-contrast bone structures but only around 52\% for low-contrast cartilage. This performance discrepancy highlights the challenges posed by imaging contrast to segmentation, while the method still achieved performance comparable to fully supervised nnU-Net in several metrics using only 5 weakly labelled reference images with point prompts.

The evaluation of reliability depends not only on peak performance but also on the stability of the model when responding to fluctuations in prompt quality. RoBox-SAM \cite{huang2024robust} effectively overcame interference caused by biological variations; even with extreme bounding box offsets of 10\% to 30\%, the model maintained a Dice score of 84.19\% in CT tasks, with its prompt robustness metric reaching levels as high as 91.89\%. This indicates that the optimized model can accurately lock onto anatomical boundaries even when manual prompts are imprecise.  Concurrently, SAM-U \cite{deng2023sam} provides a quantitative basis for clinical reliability through the expected calibration error (ECE) metric; in low-quality imaging environments, its ECE was reduced from 0.266 in the vanilla SAM to 0.015. This improvement effectively transforms prompt ambiguity into risk warnings, ensuring that the model's predictive confidence highly matches its actual accuracy.

Furthermore, maintaining physiological fidelity requires the model to transcend geometric segmentation and move toward the calculation of clinical functional metrics. The performance of LV-SAM \cite{Wu2025} in dynamic ultrasound imaging exemplifies this; the method achieved a Dice score of 94.20\%, and more importantly, a mean absolute error (MAE) of only 5.016 for the estimation of left ventricular ejection fraction.  This high consistency between pixel-level overlap and clinical functional parameters quantitatively proves that SAM, after incorporating spatio-temporal memory mechanisms, possesses the ability to capture the inertial priors of cardiac motion, thereby maintaining extremely high physiological fidelity during dynamic monitoring.\\

\noindent \textbf{Knowledge-Driven Generalization and Domain Adaptation}\\
\noindent In medical artificial intelligence, the prohibitive cost of expert annotation and the extreme heterogeneity of lesion morphologies necessitate that models possess robust generalization capabilities to bridge data distribution gaps. The core evolution of SAM in this dimension marks a paradigm shift from purely geometric-driven to knowledge-driven mechanisms. By transforming medical domain priors, such as textual descriptions, prototypical features, and spatio-temporal inertia, into prompt embeddings, SAM can precisely define generalization boundaries even with scarce data.

Experimental data proves that by relying solely on minimal and simplistic guidance, SAM can achieve high-precision segmentation in complex medical tasks, thereby fundamentally reducing the reliance on costly medical annotations. In the dense nuclei segmentation task on the MoNuSeg dataset, All-in-SAM \cite{cui2024all} achieved Dice scores of 78.16\% with complete annotations and 78.73\% with weak annotations under the extreme setting of using only 0.5\% of the training data (just 3 patches). It thus considerably outperformed the traditional fully supervised model, nnU-Net, which exhibited substantial degradation in the same scenario. This boundary was further extended by a method developed by Wu~\emph{et al.}~\cite{wu2023self} that required only 20 training images to achieve Dice scores of 62.78\% and 66.78\% on the gastrointestinal (Kvasir-SEG) and skin disease (ISIC-2018) datasets, respectively, surpassing MedSAM under identical constraints. Ablation studies further revealed the importance of knowledge density: the performance of combined point and box prompts (66.78\%) was far superior to that of point-only (61.23\%) and box-only (47.99\%) prompts, proving that hybrid geometric knowledge is critical for establishing cross-domain recognition boundaries.  

To address semantic ambiguity caused by high lesion heterogeneity, contrastive prototype learning has demonstrated exceptional semantic alignment capabilities. In unsupervised brain tumor segmentation on the BraTS2020 dataset, CLISC \cite{ma2025clisc} utilized a contrastive prototype alignment strategy to obtain a Dice score of 85.60\%, marking an improvement of over 10\% compared to other state-of-the-art unsupervised methods. Even when facing tiny lesions, its Dice score remained at 79.67\%, demonstrating that constructing contrastive constraints at the semantic level allows SAM to effectively overcome the blurred spatial boundaries caused by tumor infiltration.  Furthermore, for biological tissues with complex topological structures, BioSAM \cite{10705688} leveraged domain-specific fine-tuning in electron microscopy neuron segmentation to lower the professional metrics ``variation of information’’ (VOI) and ``adapted random error’’ (ARAND) to leading levels of 0.7799 and 0.1797, respectively. This achievement proves SAM's potential to generalize from segmenting ordinary objects to understanding complex biological connectivity. 

Finally, textual semantic priors have played a pivotal role in compensating for the lack of visual features in low-quality images. In pancreatic EUS tasks characterized by extremely low imaging contrast, TextSAM-EUS \cite{11375435} introduced textual semantic embeddings based on BiomedCLIP. Professional descriptors such as pancreatic tumor with irregular boundaries helped the model achieve a Dice of 82.69\% in a fully automated segmentation mode. Given that the model’s trainable parameter count is only 1.69 million (merely 0.86\% of the total), these results quantitatively demonstrate that the medical knowledge embedded in language models can effectively hedge against visual interference from ultrasound artifacts, enabling SAM to lock onto targets through semantic guidance even when visual information is fragmented.\\

\noindent \textbf{Clinical Utility and Computational Trade-offs}\\
\noindent The ultimate marginal benefit of clinical implementation depends not only on the absolute accuracy of the algorithm but also on the considerable reduction in human-computer collaboration costs. The evaluation dimensions of medical segmentation are shifting from a single focus on memory footprint to a comprehensive balance of click counts, fault tolerance, and inference speed. Through efficient parameter-efficient fine-tuning (PEFT) and intelligent prompting, SAM has achieved the performance necessary for near-real-time clinical deployment.

Instant adaptation capability is a core indicator for measuring clinical utility, especially in emergency or intraoperative scenarios where the fine-tuning efficiency of the model is critical. After introducing a self-prompting mechanism, Wu~\emph{et al.}~\cite{wu2023self} reduced the training time to less than 30 seconds, considerably improving efficiency compared to traditional fine-tuning methods that often take more than 30 minutes. Under this extreme time reduction, the model still maintains Dice scores of 62.78\% and 66.78\% on the Kvasir-SEG (polyps) and ISIC (skin) datasets, respectively. This balance between performance and speed showcases the immense potential of SAM, powered by prompt engineering, in medical clinical environments requiring rapid on-site feedback.

The reduction in interaction costs directly reflects the liberation of clinicians from high-intensity labor. Reinforcement learning agents have shown excellence in optimizing interaction paths. Research by Huang~\emph{et al.}~\cite{Hua_Optimizing_MICCAI2024} and Wu~\emph{et al.}~\cite{wu2023self} indicates that by simulating expert clicking logic, models can compress the average number of interaction steps required for segmentation tasks to approximately 6 steps, increasing overall interaction speed by 10 times. As a result, when processing large-scale clinical slices, doctors can obtain highly reliable segmentation results with minimal manual intervention, achieving considerable marginal benefit gains. Furthermore, lightweight architectures such as Swin-LiteMedSAM \cite{10.1007/978-3-031-81854-7_5} further lower hardware requirements through hierarchical window mechanisms, considerably enhancing deployment feasibility on mobile devices while maintaining clinically acceptable precision.

Clinical reliability is also reflected in the tolerance for human operational deviations, which is the ability to maintain stable performance under non-ideal intervention. The robustness of RoBox-SAM \cite{huang2024robust} was exhaustively tested on a multi-modal dataset containing 99k images. Even in extreme cases where the bounding box generated random offsets of 10\% to 30\%, the model still maintained a Dice score of 84.19\% in CT tasks, with its prompt robustness reaching levels as high as 91.89\%. Therefore, the model possesses a strong self-rectification capability when facing accidental operational errors or imprecise clicks from doctors, thereby safeguarding the security baseline of clinical diagnosis.

\subsection{Remote Sensing Interpretation}
\subsubsection{Domain Challenges and Adaptation Logic of Prompting Strategies}
\label{sec:domain challenges-rs}

\begin{table*}[tbp]
\centering
\scriptsize
\caption{Domain-specific adaptation strategies: Addressing remote sensing challenges through prompt engineering.}
\renewcommand{\arraystretch}{1.4}
\begin{tabular}{>{\raggedright\arraybackslash}p{2.3cm}>{\raggedright\arraybackslash}p{2.5cm}>{\raggedright\arraybackslash}p{3.25cm}>{\raggedright\arraybackslash}p{3.0cm}>{\raggedright\arraybackslash}p{1.9cm}}
\hline
\textbf{Method} & \textbf{Pub.} & \textbf{Core Challenge} & \textbf{Prompt Generation} & \textbf{Prompt Type} \\
\hline
\rowcolor{ourrow}
\multicolumn{5}{c}{\textbf{Automated Semantic Referral}} \\
\hline
Diab~\emph{et al.} \cite{diab2025optimizing} & AI Geosci. 2025 & Interaction redundancy, noisy bounding boxes & Grounding DINO with Outlier Rejection & Box \\\hline
RSTG-SAM \cite{yuan2025rstg} & IEEE GRSL 2025 & Cross-modal alignment, dense small targets & 2-stage Transformer Generator & Text, Box \\\hline
AerOSeg \cite{dutta2025aeroseg} & CVPRW 2025 & Open-vocabulary, scale/rotation variance & RS-specific Text Templates & Text \\
\hline
\rowcolor{ourrow}
\multicolumn{5}{c}{\textbf{Spatio-temporal Monitoring}} \\
\hline
SAM-CD \cite{altam2023samcd} & NeurIPSW 2023 & Bi-temporal inconsistency, blurry edges & Change Prompter from Embedding & Implicit emb. \\\hline
PUCD \cite{oh2023prototype} & arXiv 2023 & Unsupervised detection, parameter sensitivity & Prototype-oriented Synthesis & Mask\\\hline
DED-SAM \cite{qiu2024ded} & IEEE JSTARS 2024 & Boundary precision, semantic artifacts & Coarse-to-fine DED, PEFT & Mask, Box \\\hline
APSAM \cite{wang2025auto} & IEEE GRSL 2025 & Weak supervision, irregular landslide shapes & CAM-based Centroid Extraction & Point, Box \\
\hline
\rowcolor{ourrow}
\multicolumn{5}{c}{\textbf{Panoptic and Multi-scale Processing}} \\
\hline
RSPS-SAM \cite{liu2024rsps} & Remote Sens. 2024 & Global info loss, extreme scale imbalance & Batch Attention Pyramid & Mask \\\hline
InstructSAM \cite{zheng2025instructsam} & NeurIPS 2025 & Count-aware detection, tiny object recall & LVLM-guided Constraint Matching & Text \\
\hline
\end{tabular}
\label{tab:sam_taxonomy_remote_sensing}
\end{table*}

In remote sensing (RS), the adaptation logic of prompt engineering aims to bridge the gap between the encyclopedic knowledge of general-purpose foundation models and the highly specialized, complex backgrounds of Earth observation tasks. This evolution is driven not only by the need for enhanced segmentation precision but, more importantly, by the ambition to transform SAM from a passive image processing tool into an intelligent interpretation platform capable of understanding geographic semantics, spatio-temporal dynamics, and scaling laws. 

The fundamental driver of this paradigm shift lies in the inherent contradiction between general visual priors and RS imaging mechanisms: the design logic of the native SAM is primarily based on discrete objects with clear semantic boundaries, whereas RS interpretation faces unique challenges such as extreme scale imbalance under nadir-view perspectives, dynamic consistency across multi-temporal imagery, and inefficient interaction with massive geospatial targets. This misalignment in physical and semantic properties leads to 3 core problems. First, the extremely dense distribution and dramatic scale variance of geospatial objects (ranging from tiny vehicles to vast airport runways) result in considerable interaction redundancy and a high risk of missed detections for traditional manual point prompts in large-scale interpretation tasks. Second, the complex background textures of RS imagery—such as vegetation shadows and terrain illumination interference—introduce a pronounced domain gap, causing fixed prompts to suffer from severe semantic drift in zero-shot settings and leading to high false alarm rates. Finally, for dynamic monitoring requirements, the static logic of single-image segmentation fails to satisfy the rigorous demands for boundary consistency and temporal correlation in bi-temporal change detection. Therefore, as shown in Tab. \ref{tab:sam_taxonomy_remote_sensing}, prompt engineering in the RS domain is no longer a mere segmentation aid but can embed geospatial priors and spatio-temporal knowledge into models, thereby enhancing system perceptual depth.

First, the evolution from manual prompting to automated semantic referral effectively addresses the heavy dependency of RS interpretation on prompt information. To tackle the challenges of extremely dense targets and the impracticality of manual annotation in RS imagery, RSTG-SAM \cite{yuan2025rstg} and Diab \emph{et al.} \cite{diab2025optimizing} have established comprehensive semantic transformation mechanisms to achieve automated semantic retrieval. In referring expression segmentation tasks, the system automatically converts natural language instructions (e.g., ``the blue cargo ship near the pier’’) into spatial bounding box prompts. Within this workflow, Grounding DINO serves as the starting point for semantic understanding by translating abstract language into geospatial bounding boxes, while RSTG-SAM achieves a similar coarse localization function through its transformer-based prompt generator. To further enhance model adaptability to RS scenes, AerOSeg \cite{dutta2025aeroseg} introduces RS-specific prompt templates to generate text embeddings, leveraging the open-vocabulary capabilities of CLIP to improve generalization on unseen categories. These advancements collectively support a core contention: by constructing a closed-loop text-box-mask workflow, abstract semantic intent can be effectively mapped to precise pixel-level localization, liberating SAM from reliance on intensive manual point-selection and enabling automated geospatial feature extraction.

Second, the extension from static single-image segmentation to bi-temporal change detection addresses the challenges of contextual consistency through innovative prompting mechanisms, thereby bridging the domain gap inherent in RS. Addressing the artifacts caused by the complex textures of RS imagery, SAM-CD \cite{altam2023samcd} innovatively proposes a change-aware mechanism that utilizes a change modeler to extract difference embeddings from bi-temporal pairs, which are then converted into SAM-intelligible prompt information. Similarly, PUCD \cite{oh2023prototype} adopts a prototype-oriented change synthesis logic, employing the foundation model DINOv2 to extract features for unsupervised clustering, followed by SAM for edge refinement. DED-SAM \cite{qiu2024ded} introduces a hybrid coarse-to-fine architecture, utilizing traditional networks to locate change regions and generate coarse masks as prompts, then exploiting SAM’s geometric sensitivity for boundary extraction. Furthermore, in data-constrained scenarios, APSAM \cite{wang2025auto} demonstrates the potential of weakly supervised adaptation by adaptively generating point and box prompts via CAM, guiding the model to accurately extract irregular features like landslides amidst complex backgrounds. This strategy of employing SAM as a boundary refiner has become a primary pathway for narrowing the gap between general segmentation models and RS change detection tasks.

Finally, the synergistic processing of panoptic and multi-scale targets has become a critical strategy for addressing the extreme scale variations in RS scenes. Across the dramatic range of scales, from microscopic aircraft to expansive runways, dynamic prompting strategies with scale-awareness play a vital role in interpretation. RSPS-SAM \cite{liu2024rsps} introduces a batch attention pyramid module to achieve multi-scale feature extraction and capture long-range contextual information, effectively resolving the loss of global information caused by image cropping in RS processing. Its mask decoder automatically generates mask prompts and predicts categories, achieving fully automated RS panoptic segmentation. Correspondingly, InstructSAM \cite{zheng2025instructsam} approaches the problem through counting and spatial constraints, decomposing the task into instruction parsing, mask proposal, and constraint matching. Especially when handling tiny targets, it runs an additional prompt generation process on original image crops, considerably improving the recall rate of small objects and balancing precision and recall by integrating quantity constraints into the matching logic. Therefore, by designing scale-aware dynamic prompting strategies, models can better balance global context with local details, achieving high-quality intelligent interpretation of complex RS panoptic scenes.

\subsubsection{Quantitative Benchmarking and Computational Efficiency Analysis}
This section further validates the efficacy of various prompting strategies through quantitative benchmarking. As summarized in Tab. \ref{tab:sam_rs_performance}, representative models demonstrate considerable performance variances across core benchmarks, including semantic segmentation, change detection, and panoptic interpretation, reflecting their diverse adaptation pathways to geospatial challenges.\\

\begin{table*}[tbp]
\centering
\scriptsize
\caption{Quantitative benchmarking of SAM-based methods: Comparative performance across remote sensing tasks and learning paradigms.}
\renewcommand{\arraystretch}{1.6}
\begin{tabular}{>{\raggedright\arraybackslash}p{2.6cm}>{\raggedright\arraybackslash}p{3.9cm} >{\raggedright\arraybackslash}p{5.15cm}}
\hline
\textbf{Method}& \textbf{Benchmark} & \textbf{Core Performance} \\ \hline

\rowcolor{ourrow}
\multicolumn{3}{c}{\textbf{Automated Semantic Referral}} \\ \hline
Diab~\emph{et al.} \cite{diab2025optimizing}& UAV/Aerial/Satellite & 
\makecell[l]{
Acc: 98.1-99.95\% (UAV)\\
Acc: 91-99.4\% (Airborne)\\
Acc: 80\% (Satellite)}\\\hline
RSTG-SAM \cite{yuan2025rstg} & RRSIS-D & mIoU: 67.14\%, oIoU: 78.15\% \\\hline
AerOSeg \cite{dutta2025aeroseg}& iSAID, DLRSD, OEM & \makecell[l]{s-mIoU: 62.37\%, u-mIoU: 46.25\%, \\h-mIoU: 52.78\% (Avg.)}\\ \hline

\rowcolor{ourrow}
\multicolumn{3}{c}{\textbf{Spatio-temporal Monitoring}} \\ \hline
SAM-CD \cite{altam2023samcd}& LEVIR-CD, DSIFN-CD & \makecell[l]{F1: 91.52\%, IoU: 84.38\% (LEVIR-CD)\\F1: 71.02\%, IoU: 55.64\% (DSIFN-CD)} \\\hline
PUCD \cite{oh2023prototype}& LEVIR-Extension& F1: 95.1\%, IoU: 90.8\%\\\hline
DED-SAM \cite{qiu2024ded} & LEVIR-CD, SYSU-CD, CDD & \makecell[l]{
F1: 92.00\%, mIoU: 85.11\% (LEVIR-CD)\\
F1: 84.15\%, mIoU: 72.01\% (SYSU-CD)\\
F1: 97.72\%, mIoU: 95.47\% (CDD)
}\\\hline
APSAM \cite{wang2025auto}& HongKong, Turkey& \makecell[l]{F1: 79.14\%, IoU: 65.48\% (Hongkong)\\
F1: 73.71\%, IoU: 58.37\% (Turkey)}\\ \hline

\rowcolor{ourrow}
\multicolumn{3}{c}{\textbf{Panoptic and Multi-scale Processing}} \\ \hline
RSPS-SAM \cite{liu2024rsps}& RSAPS-ASD & Panoptic Quality: 57.2\\\hline
InstructSAM \cite{zheng2025instructsam}& EarthInstruct &  \makecell[l]{
IoU: 20.2\% (DIOR)\\
IoU: 14.8\% (NWPU-VHR-10)
} \\ \hline
\end{tabular}
\label{tab:sam_rs_performance}
\end{table*}

\noindent \textbf{Performance Deconstruction: Architectural Differences and Task Characteristics}\\
\noindent Quantitative experimental data reveal that the performance of SAM in remote sensing tasks is highly dependent on the precise alignment between its prompting mechanism and geospatial features, while different architectural designs fundamentally determine the depth and effectiveness.

At the geometric error level, SAM's geometric modeling is constrained by the smoothing effect of the ViT encoder on high-frequency edge information. When the edge information falls below the perceptual threshold of the ViT, its geometric modeling capability considerably degrades. Experiments by Diab~\emph{et al.} \cite{diab2025optimizing} show that on unmanned aerial vehicle (UAV) imagery (centimeter-level resolution), SAM can achieve a segmentation accuracy of 99.95\%; whereas on satellite imagery (meter-level resolution), affected by mixed pixels and edge blur, the accuracy plummets to 80\%.

At the semantic error level, the failure of text-visual alignment constitutes another critical source of error, particularly in complex remote sensing scenes. In such scenes, semantic ambiguity is especially prominent—for example, white vehicles may be confused with light-colored roofs under low resolution. RSTG-SAM \cite{yuan2025rstg} addresses this issue through a 2-stage cross-modal alignment mechanism: the first stage employs a multimodal encoder to generate text-guided sparse prompt boxes; the second stage injects semantic information into the visual feature extraction process via the TG-Adapter. This ``coarse localization to fine segmentation’’ architectural design enables the model to achieve 67.14\% mean IoU (mIoU) on the RRSIS-D dataset, representing an improvement of approximately 1.11\% compared to single-stage methods, and underscores the crucial role of semantic guidance in suppressing visual confusion.

At the structural error level, the segmentation dilemma for densely packed small targets stems from the locality of the attention mechanism and insufficient semantic correlation. This leads to phenomena where the native SAM is prone to over-segmentation (splitting a single target into multiple fragments) or under-segmentation (completely missing small targets) when processing high-density small objects. To address this challenge, RSTG-SAM \cite{yuan2025rstg} introduces a text-guided attention mechanism, enabling the model to focus on semantically relevant regions. Consequently, its segmentation accuracy for small target categories such as vehicles improves by 2.78\% compared to the strongest baseline, MAFN. As a result, structured prompting mechanisms can effectively mitigate over- and under-segmentation issues, thereby improving the model's adaptability to complex scenes.\\

\noindent \textbf{Systematic Mapping of Task Generalization Boundaries}\\
\noindent The core contradiction in remote sensing intelligent interpretation lies in the tension between massive amounts of imagery data and extremely limited pixel-level annotation resources. The co-design of prompting strategies and model architectures provides a systematic pathway to resolve this contradiction. The systematic mapping of task generalization boundaries reveals that different technical approaches exhibit clear differentiation and complementarity along 2 key dimensions: open-vocabulary recognition capability and annotation cost control. AerOSeg \cite{dutta2025aeroseg} constructs a domain-adaptive prompt template system to directionally guide the open-vocabulary capability of CLIP into the remote sensing semantic space, thereby suppressing cross-domain drift. To enhance visual-semantic alignment, spatial priors generated by SAM are further integrated as structural constraints. This forms a closed-loop optimization mechanism in which text prompts guide semantic focus, while visual masks constrain spatial scope, achieving an average harmonic mIoU (h-mIoU) of 52.78\% on benchmark datasets such as iSAID and DLRSD. As a result, through sophisticated prompt engineering, a model's semantic boundaries can be systematically expanded without relying on annotation data for unseen classes, realizing a paradigm shift from closed-set to open-set segmentation.

Unlike AerOSeg's focus on the horizontal expansion of semantic generalization capability, APSAM \cite{wang2025auto} is dedicated to pushing the potential of weakly supervised learning to its extreme in the dimension of annotation efficiency. APSAM abandons the traditional complex iterative approach aimed at generating high-precision CAM. Instead, it employs heuristic rules to transform the coarse-grained semantic information embedded in image-level labels into sparse spatial prompts, such as representative points or bounding boxes within salient regions. On the landslide extraction task, APSAM achieves an F1 score of 73.71\%, outperforming many fully supervised methods. It successfully transforms pixel-level annotation into an automatically generated intermediate product, reducing annotation labor costs by nearly 2 orders of magnitude, and provides a feasible path for resource-scarce scenarios like emergency response and global monitoring.

There exists an adaptive boundary for tasks within the current field of remote sensing prompt learning. Facing the challenge of identifying unknown categories in an open world, the semantic enhancement approach represented by AerOSeg provides a solid theoretical framework. In industrial scenarios with limited annotation resources and a need for rapid deployment, the automatic prompt generation approach represented by APSAM demonstrates higher practical efficiency. Future work could integrate the automatic prompt mechanism with open-vocabulary classifiers to build a unified system for open-vocabulary segmentation driven by extremely weak supervision. Essentially, the mapping of task generalization boundaries indicates that there is no universally optimal prompting strategy. The selection process is, in essence, a multi-objective Pareto decision-making process based on task semantic complexity, the availability of annotation resources, and computational constraints.\\

\noindent \textbf{Marginal Benefit of Computational Resources}\\
\noindent In advancing remote sensing intelligent interpretation towards practical application, computational efficiency and resource overhead have become model deployment bottlenecks. High-resolution remote sensing imagery often possesses an extremely large spatial scale (e.g., exceeding 4,000$\times$4,000 pixels), posing severe challenges to model inference speed, memory footprint, and energy consumption. Current research indicates that combining a foundation model, PEFT, and architectural innovation at the inference stage can optimize the balance between performance and efficiency.

During training, PEFT strategies demonstrate their fundamental advantage in controlling computational costs. RSTG-SAM \cite{yuan2025rstg} achieves targeted activation of the foundation model by inserting lightweight text-guided adapter modules into the frozen SAM visual encoder. Its total parameter count is 483 million, but only 174 million parameters (approximately 36\%) require updating during actual training. This partial fine-tuning strategy not only substantially reduces the graphics processing unit’s memory consumption and computation time required for training but also effectively preserves the generic visual representation capabilities that SAM acquired from billions of natural images, preventing catastrophic forgetting on data-limited remote sensing tasks. DED-SAM \cite{qiu2024ded} adopts another externally guided PEFT approach, using an independently designed change feature fusion network to generate high-quality prompts, which then drive the frozen SAM 2 model to perform fine segmentation. This design decouples change understanding and mask generation, avoiding the direct fine-tuning of SAM 2's tens of billions of parameters. Consequently, the model can adapt flexibly to change detection tasks with different characteristics at a minimal training cost.

At the inference stage, efficiency gains stem from the vertical decoupling of the computational pipeline and the horizontal optimization of memory access. InstructSAM \cite{zheng2025instructsam} decomposes the visual-language task into 2 stages: a large language model first parses the instruction and outputs structured semantic descriptions, then SAM 2 executes mask generation. This avoids the drastic increase in sequence length and computational redundancy caused by the language model directly outputting pixel information. This method reduces output tokens by 89\%, decreases overall runtime by 32\%, and renders inference time nearly independent of the number of targets, laying the groundwork for real-time interactive remote sensing interpretation. RSPS-SAM \cite{liu2024rsps} addresses the memory bottleneck in global context modeling for high-resolution images by proposing a batch attention pyramid scheme. This mechanism segments large images (e.g., images of 4,000$\times$4,000 pixels) into overlapping patches and organizes batched attention computations to achieve long-range dependency modeling across patches, without requiring entire images to be loaded into the graphics processing unit’s memory. It successfully circumvents the quadratic memory growth problem associated with transformer attention. While maintaining a high panoramic quality score of 57.2, it provides a viable throughput optimization path for large-scale geospatial mapping.

\subsection{Crack and Industrial Anomaly Detection}
\subsubsection{Domain Challenges and Adaptation Logic of Prompting Strategies}
\label{domain challenges}

\begin{table*}[tbp]
\centering
\scriptsize
\caption{Domain-specific adaptation strategies: Addressing industrial anomaly detection challenges through prompt engineering.}
\renewcommand{\arraystretch}{1.2}
\begin{tabular}{>{\raggedright\arraybackslash}p{2.65cm}>{\raggedright\arraybackslash}p{2.3cm}>{\raggedright\arraybackslash}p{3cm}>{\raggedright\arraybackslash}p{2.9cm}>{\raggedright\arraybackslash}p{2.4cm}}
\hline
\textbf{Method} & \textbf{Pub.} & \textbf{Core Challenge} & \textbf{Prompt Generation} & \textbf{Prompt Type} \\
\hline
\rowcolor{ourrow}
\multicolumn{5}{c}{\textbf{Infrastructure Inspection (Crack)}} \\
\hline
CrackESS \cite{wang2025crackess} & IEEE ITSC 2025 & Limited connectivity, edge-device speed & YOLOv8 detection & Box \\\hline
Deng~\emph{et al.} \cite{DENG2026100019} & COMPUT-AIDED CIV INF 2026 & 3D quantification, low contrast & Skeleton extraction & Point \\
\hline
\rowcolor{ourrow}
\multicolumn{5}{c}{\textbf{Zero-shot Anomaly Detection}} \\
\hline
ClipSAM \cite{LI2025129122} & IJON 2025 & Local precision, mask redundancy& CLIP segmentation & Multimodal \\\hline
SAA+ \cite{Cao_2025} & IEEE TCYB 2025 & Semantic ambiguity, high false positives& Expert rules, saliency & Multimodal \\\hline
IAP-AS \cite{10.1007/978-981-96-8186-0_6} & PAKDD 2025 & Dynamic context, unseen defects & LLM, Grounding DINO & Text \\
\hline
\rowcolor{ourrow}
\multicolumn{5}{c}{\textbf{Few-shot and Continual Adaptation}} \\
\hline
SPT \cite{10.1609/aaai.v39i12.33420} & AAAI 2025 & Domain shift, irregular shapes & Self-perception draft & Point, Box, Mask \\\hline
SAID \cite{huang2025said} & Sensors 2025 & One-shot generalization, no human prompt& Scene encoding & Implicit emb. \\\hline
UCAD \cite{liu2024unsupervised} & AAAI 2024 & Catastrophic forgetting, continuous streams & Contrastive learning & Implicit emb. \\
\hline
\end{tabular}
\label{tab:sam_taxonomy_industrial}
\end{table*}

In the field of industrial inspection and structural health monitoring, prompt engineering has evolved from a mere external aid to a core perceptual engine deeply coupled with task characteristics. The fundamental driver of this paradigm shift lies in the inherent contradiction between general-purpose large models and industry-specific requirements: the design logic of the native SAM is primarily based on discrete objects with clear semantic boundaries, while industrial defects such as micro-cracks and metal scratches often exhibit continuous field features characterized by thin topological structures, multiple branches, and high fusion with background textures. This misalignment in physical properties leads to 3 key problems. First, extreme geometric irregularity causes traditional point prompts to easily suffer from mask fragmentation when handling thin targets, failing to maintain critical topological connectivity. Second, low-contrast environments caused by bridge pier textures, lighting shadows, or corrosion introduce considerable semantic noise, leading to severe overflow in edge recognition with simple box prompts. Finally, in zero-shot settings, fixed text prompts face severe semantic drift when confronted with morphologically diverse defects, resulting in high false alarm rates. Therefore, in industrial applications, prompt engineering has evolved beyond mere segmentation. It now serves as a key method for embedding expert knowledge into models and enhancing system robustness. To systematically characterize this evolution, we categorize representative works into 3 primary task-driven paradigms based on their prompting mechanisms and addressed challenges, as summarized in Tab. \ref{tab:sam_taxonomy_industrial}.

To overcome these challenges, prompt engineering has evolved from external intervention to a closed loop of self-perception, multimodality, and quantification orientation. In terms of morphological adaptability, research focus has shifted toward constructing internal feedback mechanisms to compensate for the insufficient representational capacity of native prompts for complex geometric features. For example, SPT \cite{10.1609/aaai.v39i12.33420} introduced a draft-refine autoregressive prompt logic, using initially generated coarse masks as high-level semantic constraints and enhancing logical relationships between regions through the visual relation-aware adapter, thereby effectively repairing the fracture issues in thin cracks. Correspondingly, CrackESS \cite{wang2025crackess} automates spatial anchors by converting bounding boxes generated in real-time by YOLOv8 into rigid spatial constraints. This transforms prompts from the randomness of manual interaction to the determinism of spatial representation, ensuring precise targeting of fine structures even on resource-constrained edge devices.

In the dimension of annotation efficiency and task expansion, prompt engineering is transitioning from visual perception to physical dimensional quantification through cross-modal knowledge fusion. Addressing the common cold-start and sample scarcity challenges in industrial settings, IAP-AS \cite{10.1007/978-981-96-8186-0_6} and SAA+ \cite{Cao_2025} leverage LLMs to dynamically transform abstract physical property descriptions such as roughness and morphological proportion into precise spatial prompts, achieving compensation for missing visual features through semantic priors. For one-shot and continual learning scenarios, SAID \cite{huang2025said} and UCAD \cite{liu2024unsupervised} enable prompts to possess cross-task continuous adaptive capabilities by constructing implicit embeddings and memory banks. More critically, prompt engineering is further supporting the quantitative precision required for industrial decision-making. Deng~\emph{et al.} \cite{DENG2026100019} constructed a complete closed loop (semantic prompt, modality alignment, physical quantification) using SAM-generated 2D masks as geometric constraints to guide the back-projection of 3D point clouds, achieving sub-millimeter measurement accuracy. This evolutionary path indicates that prompt engineering is no longer merely a segmentation tool but a bridge that converts specific texture descriptions, geometric proportions, and scene prototypes into model-understandable semantics, supporting SAM in rapidly adapting to diverse industrial production lines and complex field monitoring environments without large-scale retraining.

\begin{table*}[tbp]
\centering
\scriptsize
\caption{Quantitative benchmarking of SAM-based methods: Comparative performance across industrial inspection tasks and learning paradigms.}
\renewcommand{\arraystretch}{1.4}
\begin{tabular}{>{\raggedright\arraybackslash}p{2.8cm} >{\raggedright\arraybackslash}p{4.4cm} >{\raggedright\arraybackslash}p{4.6cm}}
\hline
\textbf{Method}& \textbf{Benchmark} & \textbf{Core Performance} \\ \hline

\rowcolor{ourrow}
\multicolumn{3}{c}{\textbf{Infrastructure and Crack Segmentation}} \\ \hline
CrackESS \cite{wang2025crackess}& Khanhha's Test Set  & mIoU: 56.85\% \\ \hline
Deng \emph{et al.} \cite{DENG2026100019}& Self-constructed concrete dataset& mIoU: 62.25\% (110-shot) \\ \hline

\rowcolor{ourrow}
\multicolumn{3}{c}{\textbf{Zero-shot Anomaly Detection}} \\ \hline
ClipSAM \cite{LI2025129122}& MVTec AD & AUROC: 92.3\% \\ \hline
SAA+ \cite{Cao_2025}& MVTec AD, VisA, KSDD2, MTD & max-F1-pixel: 34.85\% \\\hline 
IAP-AS \cite{10.1007/978-981-96-8186-0_6}& MVTec AD& AP: 31.43\%, max-F1: 40.13\% \\ \hline

\rowcolor{ourrow}
\multicolumn{3}{c}{\textbf{Few-shot and Continual Adaptation}} \\ \hline
SPT-SAM \cite{10.1609/aaai.v39i12.33420}& MVTec AD, VisA & mIoU: 68.5 (avg) \\ \hline
SAID \cite{huang2025said}& Industrial-5i & mIoU: 29.96\% (one-shot) \\ \hline
UCAD \cite{liu2024unsupervised}& MVTec AD, VisA & I-AUROC: 93.0\%, P-AUROC: 0.456 \\ \hline
\end{tabular}
\label{tab:sam_performance_comparison}
\end{table*}

\subsubsection{Quantitative Benchmarking and Computational Efficiency Analysis}
In industrial anomaly detection, beyond the widely used mIoU measuring the spatial overlap between predicted masks and ground truth, image-level area under the curve (I-AUC), pixel-level area under the curve (P-AUC), maximum F1 score (F1-max), and average precision (AP) are also frequently used. I-AUC evaluates the ability to distinguish between defective and normal images, while P-AUC focuses on localization precision. Given that industrial defects often occupy a negligible fraction of the image, standard accuracy can be easily inflated by the vast number of normal pixels. Therefore, researchers lean towards F1-max and AP, as they provide a more objective measure of a model's detection ceiling and its ability to balance precision and recall under extreme class imbalance.

The considerable diversity in industrial inspection tasks leads to a localized selection of metrics, making direct numerical comparisons across papers challenging. As summarized in Tab. \ref{tab:sam_performance_comparison}, an effective evaluation strategy involves observing the performance margin that each model achieves relative to the vanilla SAM or other baseline methods. In infrastructure monitoring, the geometric continuity of cracks is the primary challenge. Studies such as Wang \emph{et al.} \cite{wang2025crackess} and Deng \emph{et al.} \cite{DENG2026100019} demonstrate that, despite using different datasets, the introduction of auto-generated box or skeleton-point prompts allows the mIoU for crack segmentation to more than double compared to baselines. Therefore, strong spatial constraints are essential for compensating for the general-purpose model's inherent perceptual deficiencies when handling thin, low-contrast industrial targets.

In contrast, anomaly detection models for precision quality control follow a different evolutionary path. Methods represented by SPT-SAM \cite{10.1609/aaai.v39i12.33420} and UCAD \cite{liu2024unsupervised} explore the upper bounds of precision under the self-tuning or unsupervised paradigms, achieving remarkably high I-AUC ($>93\%$--$98\%$) on standard benchmarks like MVTec AD. This suggests that when the industrial environment is relatively stable and modest fine-tuning is permitted, prompt engineering can effectively overcome the domain shift of the vanilla SAM, bringing performance close to industrial-grade robustness. Meanwhile, works such as IAP-AS \cite{10.1007/978-981-96-8186-0_6} and SAA+ \cite{Cao_2025} challenge the more rigorous zero-shot scenarios. Without access to target-domain priors, these studies utilize F1-max to validate the model's semantic perception of unseen defects. This divergence in metrics is not a matter of superiority but reflects a research shift from simple pixel coverage to sophisticated semantic precision.

While accuracy metrics are difficult to align across publications, computational efficiency serves as a concrete indicator for assessing industrial deployment potential. Experimental data indicates that the complexity of the prompt generation mechanism directly dictates real-time performance. CrackESS \cite{wang2025crackess} achieves a system throughput of 46 frames per second on the Jetson Orin Nano edge platform by converting YOLOv8 detection results into spatial prompts, a rare feat among large-model-based industrial solutions. This stands in sharp contrast to LLM-driven approaches like IAP-AS \cite{10.1007/978-981-96-8186-0_6}, which, despite their superior dynamic semantic understanding, suffer from latencies of several seconds, relegating them to offline auditing rather than online production line monitoring.

Furthermore, lightweight tuning strategies strike a balance between efficiency and precision. The results from SPT-SAM \cite{10.1609/aaai.v39i12.33420} provide an important quantitative conclusion: by adding only 0.397\% of trainable parameters, the model maintains an inference speed of approximately 7.98 frames per second on an RTX 3090. This self-perception tuning approach enhances the representation of anomaly images without incurring excessive computational overhead, representing an economically viable path for addressing domain shifts in industrial contexts.

The heterogeneous performance across various benchmarks reflects an impossible trinity in current industrial deployment: the simultaneous achievement of extreme real-time speed, zero-shot generalization, and pixel-level absolute precision. Therefore, practical applications necessitate a trade-off based on specific requirements. For tasks demanding instantaneous feedback, such as UAV-based bridge inspections, the combination of automated box prompts and a lightweight backbone is the best practice. For flexible production lines with frequent product changes, hybrid prompts based on cross-modal semantic generation maximize the benefits of zero-shot generalization. For defect segmentation of extremely irregular and fragmented shapes, the inclusion of a self-perception refinement mechanism is indispensable for ensuring detection connectivity.

\section{Challenges and Open Issues}
\subsection{Interpretability and Decision Trustworthiness}
Despite the considerable flexibility and practicality that prompt engineering brings to SAM, the black-box nature of its decision-making mechanism still severely limits its potential for deployment in high-risk, low-tolerance professional domains such as medical imaging and autonomous driving.

First, the extreme sensitivity of prompt configuration poses a fundamental challenge to system robustness. Studies like PointPrompt \cite{quesada2024benchmarking} indicate that this sensitivity is not merely a source of performance fluctuations but reflects discontinuities in geometric representation. Even minor point displacements can trigger catastrophic shifts in the generated mask. Although Stable-SAM~\cite{fan2025stable} introduces a stability metric to quantify such volatility, these approaches are essentially post-hoc analyses. The core challenge lies in embedding architectural constraints of equivariance into the inference mechanism itself, ensuring consistent outputs for semantically equivalent prompts.

More critically, this prompt sensitivity further reveals the credibility risks posed by SAM's unpredictable failure behaviors in real-world applications. In interactive settings, users often cannot anticipate under what input conditions the model might systematically fail—a concern especially pronounced in non-ideal conditions. Ji~\emph{et al.}~\cite{ji2024segment} demonstrate that SAM exhibits notable systemic biases under occlusion, low illumination, or cluttered backgrounds, consistently favoring prominent foreground objects while neglecting non-salient regions such as shadows and transparent areas. In medical imagery, this may result in missed detections of blurry lesions. Such failures cannot be easily remedied through prompt refinement, and users struggle to foresee when the model will break down, resulting in poor predictability and interpretability of model behavior.

Against this backdrop, integrating decision interpretability enhancements and uncertainty quantification mechanisms has emerged as a key direction for improving model trustworthiness. For example, MedSAM-U~\cite{11201277} visualizes model uncertainty by generating entropy maps, offering users an intuitive feedback signal regarding confidence levels. However, despite numerous prompt optimization strategies that improve robustness via geometric or semantic perspectives, there remains a lack of systematic modeling regarding how prompts modulate attention distributions within SAM and ultimately influence the mask generation pathway. While works such as ProMISe~\cite{wang2024promise} and CPC-SAM~\cite{miao2024cross} offer valuable heuristic frameworks, they largely remain empirical and fall short of revealing the causal relationships among prompt semantics, attention responses, and final outputs.

\subsection{Long-tail Distributions and Open-vocabulary Generalization}
\label{subsec:long-tail-challenges}
Despite SAM’s strong zero-shot generalization, prompt engineering still reveals a series of structural challenges when encountering long-tail category distributions and open-vocabulary semantics in real-world applications. These issues do not stem simply from insufficient data scale or model capacity, but rather from a fundamental mismatch between semantic representation, and spatial alignment.

One of the most prominent challenges in open-set segmentation is semantic drift and the failure of cross-modal alignment. In open-vocabulary scenarios, textual prompts often involve rare objects or professional terminology not covered in training corpora like SA-1B. To compensate for this semantic void, numerous works \cite{LI2025129122, vetoshkin2025talk2samtextguidedsemanticenhancement} rely on vision-language models such as CLIP to provide semantic embeddings. However, research like Open-Vocabulary SAM \cite{yuan2024open} explicitly points out that the global image-level contrastive features learned by CLIP are difficult to align effectively with the pixel-level segmentation representations of SAM. This misalignment leads to considerable performance degradation, particularly when processing small objects or rare categories. Essentially, the performance drop in these cases is a consequence of the lack of a consistent mapping mechanism between high-level semantic representations and low-level spatial structures.

The issue of false positives induced by granularity mismatch further amplifies risks in long-tail and open-vocabulary scenarios. When the prompt semantics are inherently abstract or ambiguous (e.g., ``sports equipment'' or ``decorations''), existing prompt mechanisms often fail to provide sufficient geometric and structural constraints for precise pixel-level segmentation. This frequently leads the model toward over-segmentation or forced interpretation of concepts that either do not exist or are only vaguely mentioned. Such false positive phenomena are particularly prevalent among long-tail categories, and are an unavoidable systemic problem in open-set segmentation. While SAM 3 \cite{carion2025sam} attempts to decouple ``whether a concept exists'' from ``where the object is'' by introducing a presence head—thereby suppressing some spurious detections for simple noun phrases—this mechanism primarily functions as a predictive-stage correction. It does not resolve the problem of how prompt semantics should be parsed, decomposed, and transformed into executable segmentation control signals. For complex, compositional long-tail descriptions, existing frameworks still lack a stable and reliable parsing path.

The limitations in cross-domain generalization further reveal the amplification of the aforementioned issues in professional scenarios. In fields such as remote sensing, microscopic imaging, and pathological analysis, target categories often exhibit a distinct long-tail distribution, with semantic concepts and visual patterns that differ considerably from those in natural images. In these scenarios, not only are boundaries blurred and structures complex, but the semantics themselves often lack clear correspondences in natural language, making it difficult for both geometric and semantic prompts to provide stable and effective control signals. Current prompt engineering methods in these domains often require extensive domain adaptation or additional assumptions, and their cross-domain generalization still lacks a unified, reusable supporting mechanism. This limitation suggests that the current prompt-to-segmentation paradigm, when facing the open world and long-tail semantics, remains constrained by its ability to model the relationship between high-level semantic understanding and spatial execution.

\subsection{Limitations in Complex Real-World Scenarios}
While SAM and its variants excel on standardized datasets, their performance often degrades in complex real-world settings due to occlusion, motion blur, low contrast, or cluttered backgrounds. These conditions weaken the saliency of object boundaries, making it difficult for both spatial and semantic prompts to capture accurate localization cues. Particularly when objects are occluded or partially visible, SAM tends to produce fragmented or incomplete segmentation results. Existing prompt engineering techniques are primarily designed for idealized scenarios and struggle to handle such ambiguities effectively. Additionally, multimodal prompts, such as text-visual fusion, are susceptible to semantic ambiguity in cross-domain applications. For example, the term ``tumor'' may manifest differently across medical imaging modalities, leading to localization errors. Addressing these challenges requires integrating domain-specific prior knowledge, such as topological constraints, with dynamic multi-prompt collaboration mechanisms, alongside the development of more generalizable cross-modal alignment methods.  

\subsection{Computational Efficiency and Deployment Constraints}
SAM's large size, exemplified by its ViT-H image encoder, imposes high computational demands, leading to latency issues that hinder deployment on edge devices and real-time systems. Although studies have employed techniques like model pruning, lightweight prompt tuning, and distillation to reduce computational overhead, these methods often come at the cost of segmentation accuracy. Future directions include developing more efficient prompt encoding architectures, such as sparse attention mechanisms, hierarchical prompt engineering techniques like coarse-to-fine localization, and hardware-friendly quantization and distillation techniques to balance efficiency and precision.

\subsection{Multi-Prompt Conflicts and Misalignment}
Curriculum prompting \cite{zheng2024curriculum} reveals potential conflicts between point and box prompts in staged prompting strategies. When their spatial guidance signals contradict—for instance, if a box prompt covers background regions—confusion may arise. This conflict with overlapping foreground points can degrade the model’s segmentation performance. This issue is especially prominent in scenarios with blurred boundaries or partial occlusion.  

Furthermore, inconsistencies between high-level semantic prompts and low-level geometric prompts can cause severe segmentation bias due to cross-modal misalignment. For example, in ClipSAM \cite{LI2025129122}, insufficient spatial overlap between CLIP-generated text heatmaps and detector-provided box prompts may lead to large deviations in segmentation results. Such problems, more common in open-vocabulary and zero-shot tasks, stem from the lack of unified alignment standards across text and visual embedding spaces, as well as the rigid spatial constraints of geometric prompts conflicting with the abstract nature of semantic prompts. While existing solutions, such as cross-modal attention and contrastive learning, partially mitigate this issue through feature interaction, challenges in dynamic coordination persist. For instance, semantic class cues from text prompts may override local geometric details, or fixed box ranges may suppress flexible boundary adjustments guided by text. Future work should explore finer-grained modality interaction mechanisms, such as differentiable prompt weighting, as well as joint optimization objectives, such as modality-consistency loss functions, to achieve hierarchical synergy between semantic and geometric prompts. This will ensure robust multi-prompt performance in both spatial accuracy and semantic consistency.

\section{Future Research Directions}
\subsection{Adaptive Prompt Regulation via Meta-Learning and Test-Time Adaptation}
Current SAM-based prompt engineering methods predominantly adopt static or semi-static prompt generation strategies, where the form and parameters of prompts are determined prior to inference and remain unchanged throughout the segmentation process. These designs incorporate mechanisms such as prototype learning and similarity matching, which enhance consistency across samples and tasks. However, they remain built upon predefined representational spaces and fixed structures. As a result, they struggle to cope with challenges in real-world applications, such as sparse object distributions, variable imaging conditions, and ambiguous semantic boundaries. This limitation becomes particularly pronounced under long-tailed categories, cross-domain generalization, or extreme noise conditions. In these scenarios, prompt engineering lacks the capacity to adapt dynamically to the task context. This lack of adaptability has become an increasingly critical bottleneck restricting the generalization potential of SAM.

To overcome this limitation, it is promising to introduce meta-learning and test-time adaptation mechanisms into the prompt engineering framework, thereby shifting the paradigm from static configuration to dynamic evolution. Existing methods such as SurgicalSAM and APL-SAM are essentially lookup-based prototype matching approaches that rely on pre-constructed prototype libraries for prompt generation and often falter when facing unseen categories or entirely new domains. In contrast, the meta-learning paradigm emphasizes learning how to generate and adjust prompts, moving beyond the pursuit of an optimal prompt representation for each task. Instead, it aims to acquire transferable strategies through cross-task training. For instance, MetaSPO~\cite{choi2025system} demonstrates the potential of meta-strategy learning to achieve superior transfer performance in language tasks. Similarly, in the vision domain, one can explore the construction of prototype-inductive meta-strategies, enabling the model to automatically infer the most appropriate prompting logic for the current task based on a small set of support samples. This approach eliminates the dependency on static prototypes and enables learning on the fly across tasks. Analogous to task-aware strategy learning in few-shot settings as in MetaUAS~\cite{NEURIPS2024_463a91da}, such episodic training mechanisms could empower SAM with rapid inductive capabilities in novel scenarios.

Meanwhile, introducing real-time feedback mechanisms during inference to achieve closed-loop prompt adjustment is another critical path for the evolution of prompt engineering. Most current methods employ one-shot prompt generation, lacking mechanisms to utilize intermediate predictions for iterative refinement. This results in difficulties when handling dynamic deviation or structural drift during inference. Future systems should aim to develop feedback-driven online adaptation mechanisms, such as using uncertainty estimation (e.g., entropy maps in MedSAM-U~\cite{11201277}) to guide prompt adjustments or leveraging temporal consistency modeling for geometric alignment. For example, CAV-SAM~\cite{wang2025correspondence} proposes a test-time geometric alignment module that performs lightweight fine-tuning based on a single reference image while enhancing robustness against shape deformation through a cycle consistency strategy. Viewing prompt configuration as a tunable control variable rather than a static input could enhance SAM's stability in complex conditions such as occlusion or sensor noise. This approach provides a pathway for deployment in resource-constrained and interactive segmentation scenarios.

\subsection{Synergistic Co-design of Prompt Engineering and Model Architectures}
As prompt engineering becomes increasingly prevalent in interactive image segmentation, its role is no longer confined to static input configurations external to the model. Prompt engineering should move toward a stage of co-design between prompts and model architectures and should aim to build systems that are more efficient, stable, and synergistically perceptive. Such co-design is not limited to enhancing the expressiveness of semantic injection, but also encompasses a deep integration with system-level efficiency and deployment adaptability.

First, deep semantic injection of prompts provides a critical anchor for co-design. Conventional approaches typically embed prompts at the model input or within early feature layers, given that semantic information is prone to dilution as it propagates through deeper layers, limiting the sustained influence of intent on the visual perception process. In contrast, methods such as Visual Prompt Tuning~\cite{jia2022visual} and Explicit Visual Prompting~\cite{liu2023explicit} demonstrate that prompt engineering can be introduced as intermediate-layer guidance signals embedded into the backbone in the form of dynamic biases or intent adapters. This design encourages the model to focus on prompt-indicated regions early during feature extraction. The transition from semantic supplementation to perceptual guidance redefines prompts as native control mechanisms of model behavior rather than external correction terms. Such a shift not only bolsters segmentation stability but also ensures intent consistency when dealing with complex environmental challenges.

Second, co-design should further extend to the efficiency co-optimization between prompt engineering and system architectures. Lightweight image encoders have evolved rapidly, with models like MobileSAMv2~\cite{zhang2023mobilesamv2} replacing dense grid-based sampling with high-quality bounding box prompts to cut redundant computation~\cite{wang2025crackess,10648107}. Yet, the prompt generation and processing branches remain critical in large-scale tasks such as ``segment everything.''

These bottlenecks are the result of a lack of structural co-optimization between prompt modules and backbone models. Even when the backbone model is compressed, the overall system often fails to meet the dual requirements of low latency and low power consumption for edge deployment. Looking forward, prompt engineering design should not remain isolated from system architecture. Instead, it should incorporate hardware-aware neural architecture search to jointly optimize prompt encoding protocols according to platform-specific characteristics like neural processing units and digital signal processors. In addition, joint distillation between prompt modules and backbone networks can further eliminate computational redundancy. Under the premise of preserving segmentation accuracy, such system-level co-optimization enables genuine improvements in latency and energy efficiency, and pushes prompt engineering toward deployable and practical interactive segmentation systems.

\subsection{Enhancing Prompt Robustness with Causal Inference}
Current prompt engineering strategies primarily rely on statistical correlations between prompts and segmentation performance, often failing in generalization scenarios due to their inability to reveal true causal mechanisms. Future research should integrate causal inference frameworks to distinguish key prompt features that genuinely affect segmentation accuracy from spurious correlations. For instance, causal graphs could explicitly model how prompt positions or structures influence boundary precision while accounting for confounding factors like image noise or object occlusion. Counterfactual prompt engineering could systematically alter prompt features to simulate different scenarios, evaluating their direct impact on segmentation results and revealing stable causal pathways. This approach is particularly valuable in safety-critical domains like healthcare, where anatomical variations may create misleading correlations. Combining causal discovery with prompt optimization could enable stable model performance under unknown conditions or atypical data distributions.

\subsection{Multi-Agent Collaborative Prompt Framework}
As segmentation tasks grow more complex, single-prompt generation mechanisms show limitations. A promising solution is a multi-agent system where specialized agents collaboratively optimize prompts: a spatial localization agent generates target region bounding boxes, a semantic parsing agent processes text descriptions or environmental metadata, and an uncertainty assessment agent monitors prediction confidence. These agents could interact through cross-attention mechanisms or gradient-based negotiation protocols. For example, in remote sensing, a terrain-aware agent could dynamically adjust urban detection prompts using geographic information system data. Key challenges include designing reward mechanisms that balance specialization and consensus, as well as developing efficient cross-agent knowledge transfer methods to avoid redundant computations. Such frameworks could considerably enhance adaptability in dynamic multi-domain scenarios like autonomous driving or precision agriculture.

\subsection{Progressive Prompt Generation Based on Diffusion Models}
Diffusion models offer a novel approach to prompt generation by simulating the iterative refinement process of human annotation. A diffusion-based prompt generator could start with randomly distributed point markers or coarse bounding boxes, subsequently refining them progressively through denoising steps that incorporate image features. The system could dynamically adjust its optimization path based on intermediate segmentation feedback, thereby enhancing its applicability in high-precision domains like surgical navigation or 3D medical image segmentation. Key innovations include hybrid latent space representations that combine geometric and semantic features, as well as adaptive denoising scheduling strategies to balance speed and accuracy.

\subsection{Unsupervised Prompt Adaptation Techniques}
The heavy reliance on labeled data in current methods limits applications in annotation-scarce domains. Future research should explore unsupervised or weakly supervised prompt learning techniques. For example, prompt generators could be pre-trained using proxy tasks like masked region prediction or visual-text alignment. Cross-modal distillation could leverage vision-language models like CLIP to infer relationships between image features and text descriptions such as automatically locating prompt points based on ``tumor'' mentions in radiology reports. Test-time adaptation techniques like entropy minimization or memory bank retrieval could further optimize prompts for specific images. These approaches would be invaluable in scenarios with scarce annotations, such as rare disease analysis or historical document segmentation, while maintaining strong robustness across diverse applications across increasingly diverse real-world applications.

\section{Conclusion}
We present a systematic review of prompt engineering within the Segment Anything Model (SAM), proposing a function-oriented taxonomy that categorizes prompt strategies into geometric, textual semantic, and multimodal fusion types. We further analyze the adaptation logic of these strategies across high-complexity domains such as medical imaging, remote sensing, and industrial anomaly detection. By incorporating detector-assisted prompting, prototype-based learning, and reinforcement learning-driven interaction, we demonstrate the potential of automated prompt generation to reduce annotation cost and enhance cross-domain generalization. Despite these advancements, in real-world applications, SAM still faces critical challenges, including high sensitivity to prompt quality, instability in complex environments, limited recognition of long-tail targets and rare structures, and computational inefficiencies under high-resolution or real-time constraints. Future research may address these issues through causal reasoning and uncertainty modeling to improve robustness, collaborative multi-agent prompting mechanisms for system-level coordination, and diffusion-based or progressive strategies for prompt generation and refinement. Advancing towards adaptive and systematized prompt engineering will enable SAM to achieve a better trade-off between accuracy, efficiency, and generalization, paving the way for its evolution from a foundational segmentation model to a versatile perception and execution engine across diverse application scenarios.

\section*{Acknowledgments}
\subsection*{General}
AI-assisted technologies were used as aids in the writing of the manuscript.

\subsection*{Funding}
This work was supported in part by the National Natural Science Foundation of China under Grant 62402341, in part by the Postdoctoral Fellowship Program of China Postdoctoral Science Foundation (CPSF) under Grant GZC20241225, and in part by the China Postdoctoral Science Foundation under Grant Number 2025M771513.

\subsection*{Author Contributions} 
Yidong Jiang conducted the literature review and drafted the manuscript. Jiangtong Li guided the structural design and gave critical feedback. Dawei Cheng reviewed the draft and finalized the manuscript for clarity and logical flow. All authors reviewed and approved the final version.

\subsection*{Conflicts of Interest}
The authors declare that there is no conflict of interest regarding the publication of this article.

\printbibliography
\end{document}